\documentclass[times,review,10pt]{elsarticle}  

\usepackage[utf8]{inputenc}
\usepackage[T1]{fontenc}
\usepackage{import}
\usepackage{multirow}
\usepackage{booktabs}
\usepackage{amssymb}
\usepackage{amsmath}
\usepackage{graphicx}
\usepackage[export]{adjustbox} 
\usepackage{caption}
\usepackage{float}
\usepackage{subcaption}
\usepackage{threeparttable}
\usepackage{enumitem} 
\usepackage{tikz}
\usepackage[mode=buildnew]{standalone}
\usepackage{Box}
\usepackage{ifthen}
\usepackage[pagebackref,breaklinks,colorlinks]{hyperref}
\usepackage{orcidlink}
\usepackage{setspace}

\usepackage{lmodern}
\newcommand*{\escape}[1]{\texttt{\textbackslash#1}}

\DeclareMathOperator*{\argmax}{arg\,max}
\DeclareMathOperator*{\softmax}{soft\,max}

\definecolor{Grey}{RGB}{229, 232, 232}
\definecolor{Blue}{RGB}{ 127, 179, 213 }

\usepackage{pifont}

\newcommand{\xmark}{\ding{55}}

\newcommand{\ie}{\textit{i.e.}}
\newcommand{\eg}{\textit{e.g.}}
\newcommand{\mb}[1]{\boldsymbol{#1}}

\newcommand{\mtdan}{Document Attention Network with Multi-Tokens predictions}
\newcommand{\mtdans}{MT-DAN}

\newcommand{\wdan}{Document Attention Network with Windowed queries}
\newcommand{\wdans}{W-DAN}

\newcommand{\metadan}{Meta Document Attention Network}
\newcommand{\metadans}{Meta-DAN}

\newcommand{\rotate}[1]{\rotatebox[origin=c]{90}{#1}}

\newcommand*\circled[1]{\tikz[baseline=(char.base)]{%
            \node[shape=circle,draw,inner sep=2pt] (char) {#1};}}

\journal{Pattern Recognition} 

\begin{document} 

\begin{frontmatter}


\author{Denis Coquenet\corref{cor1}}
\ead{denis.coquenet@irisa.fr}
\cortext[cor1]{Corresponding author}
\affiliation{organization={Univ Rennes, CNRS, IRISA - UMR 6074, F-35000 Rennes},
            country={France}}

\title{\metadans{}: towards an efficient prediction strategy for page-level handwritten text recognition}




\begin{abstract}
Recent advances in text recognition led to a paradigm shift for page-level recognition, from multi-step segmentation-based approaches to end-to-end attention-based ones. 
However, the naïve character-level autoregressive decoding process results in long prediction times: it requires several seconds to process a single page image on a modern GPU.
We propose the \metadan{} (\metadans{}) as a novel decoding strategy to reduce the prediction time while enabling better context modeling. 
It relies on two main components: windowed queries, to process several transformer queries altogether, enlarging the context modeling with the near future; and multi-token predictions, whose goal is to predict several tokens per query instead of only the next one.
We evaluate the proposed approach on 10 full-page handwritten datasets and demonstrate state-of-the-art results on average in terms of character error rate.
Source code and weights of trained models are available at \url{https://github.com/FactoDeepLearning/meta_dan}.
\end{abstract}




\end{frontmatter}


\section{Introduction}

Handwritten Text Recognition (HTR) is the process of automatically extracting handwritten text from a digitized image and converting it into machine-readable text.  
The historical approach to tackle this task, still in the majority today, consists of a multi-stage paradigm. 
A document image is segmented into sections of text, which are then ordered to preserve a coherent reading order. Thereafter, these text sections are recognized and reassembled to obtain the overall transcription. 
Over time, the nature of the textual section has broadened, from the isolated character~\cite{Arica2001}, to the word~\cite{ElYacoubi1999}, then to the line~\cite{Wigington2017,Michael2019,Coquenet2020} and more recently to the paragraph~\cite{Bluche2016,Bluche2017,Yousef2020,Coquenet2023a}, gradually increasing the context usable during recognition.
However, regardless of the granularity, this multi-step approach comes with significant drawbacks: 
1) the segmentation step requires an additional segmentation model and annotations, 
2) the ordering step is generally a dataset-specific hand-crafted algorithm, 
3) the recognition of the individual sections is carried out independently, limiting the context modeling during prediction, 
4) errors accumulate in each of these steps.

More recently, the Document Attention Network (DAN)~\cite{Coquenet2023b} was proposed as the first end-to-end approach for HTR at page level, achieving similar recognition performance compared to the multi-step approach. 
The DAN architecture includes a transformer-based decoder to perform a character-level autoregressive prediction process. 
Due to its state-of-the-art performance, the DAN is currently the default approach used for end-to-end handwritten text recognition at page-level~\cite{Coquenet2023c,Castro2024}. In addition, it has been extended to handle other tasks at page level such as named entity recognition~\cite{Constum2024,Constum2024b}, table recognition~\cite{Boillet2024}, and key-value information extraction~\cite{Tarride2023a,Tarride2023b}, by adding dedicated tokens in the predictable token set. 

The main drawback of the DAN, which is also an advantage, lies in its character-level iterative prediction process.
Indeed, on the one hand, processing text at character level reduces the vocabulary, making training easier. It also enables to manipulate lower-dimensional latent states, since each token carries little information, reducing computational complexity. 
On the other hand, it increases the target sequence length, and so the number of decoding iterations, increasing the prediction time. 

In this work, we propose the \metadan{} (\metadans{}), a novel prediction strategy to improve both inference speed and recognition accuracy for end-to-end page-level handwritten text recognition. 
It lies on the combined use of two distinct approaches. 
The first one is based on \textit{multi-token predictions} as it showed interesting results for Natural Language Processing (NLP)~\cite{Gloeckle2024}. 
The second consists of \textit{windowed queries}, \ie{}, processing successive transformer queries altogether as a sliding window to process them in parallel. Our main contributions are as follows:
\begin{itemize}
    \item We propose two prediction strategies to improve page-level handwritten text recognition: windowed queries and multi-token predictions.
    \item We study the introduction of a confidence-based dynamic prediction policy to strike a balance between prediction speed and recognition performance for multi-token predictions.
    \item We provide an extensive evaluation over 10 various public datasets. To our knowledge, this is the largest benchmark evaluated at page-level. This evaluation highlights that combining both proposed approaches (\metadans{}) leads to on-average state-of-the-art recognition performance. 

\end{itemize}

\section{Related Works}

\subsection{Handwritten text recognition}

State-of-the-art approaches can be divided into two main categories we refer to as the \textit{alignment-based} and the \textit{attention-based} approaches. 

The \textit{alignment-based} approaches consider the recognition task as a one-dimensional sequence alignment problem between a predicted probability lattice and a ground truth sequence. The Connectionist Temporal Classification (CTC)~\cite{Graves2006} was proposed in 2006 to perform this alignment, enabling the end-to-end recognition of word~\cite{Graves2008,Carbonell2020} and line~\cite{Wigington2017,Coquenet2020} images. 

However, the CTC only deals with one-dimensional sequences. This way, tricks have been proposed to assimilate a paragraph to a long line of text, through bilinear interpolations~\cite{Yousef2020}, or concatenation~\cite{Coquenet2021}. 
But, to our knowledge, alignment-based approaches are not suited for end-to-end full-page recognition, since the complexity of page layouts (\eg{}, multi-column pages) prevent to reduce it to a one-dimensional sequence alignment problem.

More recently, the \textit{attention-based} approaches emerged, particularly with the rise in popularity of the Transformer architecture~\cite{Vaswani2017}. 
They rely on attention mechanisms to provide an iterative autoregressive decoding process at character level, performing a kind of implicit character segmentation through the attention maps.

Contrary to the alignment-based approaches, the attention-based approaches can be applied to any input image, from a single line~\cite{Michael2019,Barrere2022} or paragraph~\cite{Bluche2017,Singh2021} to a two-page document with the DAN-based works~\cite{Coquenet2023b,Coquenet2023c,Constum2024}.

It has to be noted that marginal works~\cite{Bluche2016,Coquenet2023a} were proposed to handle paragraph recognition with a \textit{hybrid} approach combining the attention mechanisms, to generate an implicit line segmentation, and the alignment-based approach, to recognize them iteratively through CTC decoding. 
In addition, an attempt to adapt the CTC to 2D-to-1D alignment was proposed in~\cite{Schall2018}, but it did not lead to convincing results.

In summary, alignment-based approaches are advantageous because of their one-shot prediction, but they are limited to images with a single column of text, whereas attention-based approaches are capable of handling entire pages with complex layouts, but involve longer prediction times due to their autoregressive nature.

In this work, we focus on reducing the prediction time of the attention-based character-level approach to reach fast end-to-end page-level predictions, while improving the context modeling through the integration of the near future in the decoding process.

\subsection{Speeding up autoregressive processes}

The prediction time issue of autoregressive decoders is not limited to HTR. 
In~\cite{Qi2020}, a \textit{multi-token prediction} strategy was proposed to speed up text summarization: a transformer model is trained to predict the $n$ future tokens at once using $n$ projection heads. This idea was also successfully applied for neural machine translation~\cite{Stern2018} and for large language models~\cite{Gloeckle2024}.

The authors of~\cite{Fei2021} proposed another way to reduce the prediction time for the task of image captioning. The target sequence is split into subgroups of words, which are decoded in parallel. 

Closer to our work, a similar approach was proposed to speed up the Document Attention Network by splitting the decoding process into two stages~\cite{Coquenet2023c}. 
The first stage is dedicated to predicting the first character of each line, and the second one consists in completing all lines in parallel, reducing the number of decoding steps. 

The \textit{windowed queries} we propose also rely on the idea of processing queries in parallel, but they differ in that the processed queries are successive, leading to a unique decoding stage and a context mixing full past and near future.

An alternative to speed up the prediction process is to reduce the length of the target sequence, by relying on a dedicated tokenizer, such as the Byte-Pair Encoding (BPE) algorithm~\cite{Sennrich2016}.

This technique was applied in~\cite{Constum2024}, showing a significant prediction time decrease, in return for a higher character error rate. The approach we propose is compatible with this kind of technique as it relies on the token prediction strategy, regardless of the nature of the tokens.

In this paper, we propose to combine both windowed queries and multi-token predictions to speed up the prediction process while improving the context modeling capability of the model.

\section{Background}

\subsection{HTR problem formulation}
Given an input image $\mb{X}$, the goal is to extract the textual information $\mb{y}$ in a human-coherent order. 
The input image $\mb{X} \in \mathbb{R}^{H \times W \times C}$ can be of variable size with height $H$, width $W$, and encoded over $C$ channels (typically 1 for grayscale, and 3 for RGB). 
The output $\mb{y}$ corresponds to the whole textual information, \ie, all the text lines put together. We note $\mb{l}^i$ the $i$\textsuperscript{th} text line of length $L_i$. 
The text lines $\mb{l}^i$ are sequences of characters $\mb{l}^i = [\mb{l}^i_1, \ldots{}, \mb{l}^i_{L_i}]$ such that $\mb{l}^i_j \in \mathcal{A}^\text{c}$, where $\mathcal{A}^\text{c}$ is a given alphabet ($|\mathcal{A}^\text{c}|$ refers to its cardinality). 
The global output $\mb{y}$ is the concatenation of the $N$ text lines that make up the document, separated by a special join token (typically a white space or new line character). 
In this way, $\mb{y} = [\mb{l}^1_1, \ldots{}, \mb{l}^1_{L_1}, \escape{n}, \mb{l}^2_1, \ldots{}, \mb{l}^2_{L_2}, \escape{n}, \ldots{},  \mb{l}^N_1, \ldots{}, \mb{l}^N_{L_N}]$ has a length of $ \displaystyle L^y=\sum_{i=1}^{N} L_i + N - 1$ and $\mb{y}_1 = \mb{l}^1_1$.

\subsection{Document Attention Network}

\begin{figure}[ht!]
    \centering
        \includegraphics[width=\textwidth]{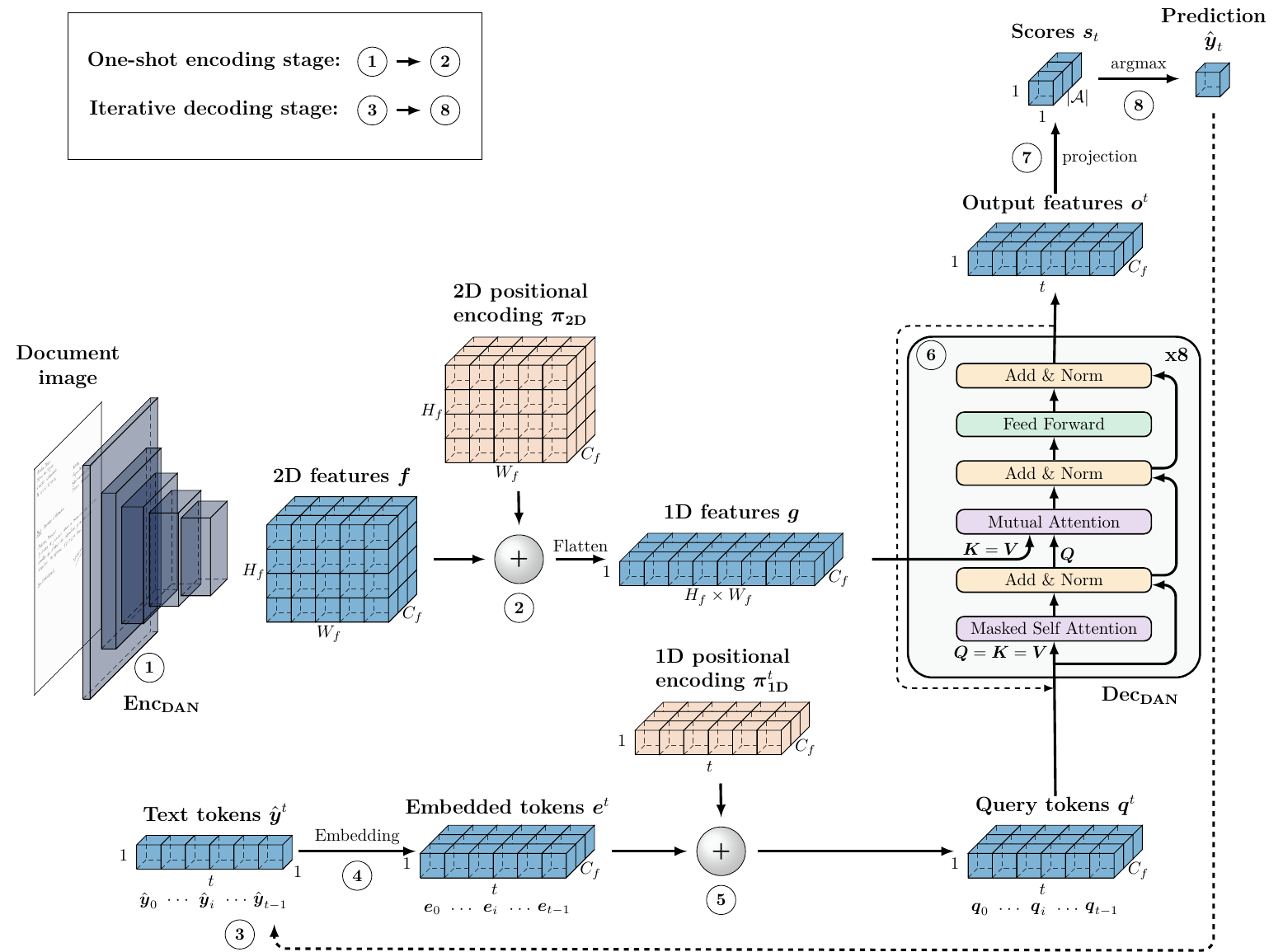}
    \caption{DAN architecture. Encoding stage: 2D features $\mb{f}$ are extracted from the input image through a Fully Convolutional Network (FCN). 
    Decoding stage: at each iteration $t$, a query sequence $\mb{q}^t$ is generated from the previous predictions $\hat{\mb{y}}^t$ and associated with the image features through the transformer's attention mechanisms to predict the next token $\hat{\mb{y}}_t$. The decoding is initialized with a start-of-prediction token $\hat{\mb{y}}_0 = \text{<s>}$ and stops when the end-of-transcription token <e> is predicted.}
    \label{fig:dan-archi}
\end{figure}

The Document Attention Network (DAN)~\cite{Coquenet2023b} was proposed to tackle this task as a character-level autoregressive prediction process. It refers to a deep neural network illustrated in Figure~\ref{fig:dan-archi}. It is made up of two main building blocks: a Fully Convolutional Network (FCN) encoder and a causal transformer-based decoder. 
The general idea of this approach is as follows. Based on the image features extracted by the encoder, and on the previously predicted tokens, the goal is to predict the next token (\ie, a character). 
To determine when to stop this iterative decoding process, a special end-of-transcription token (<e>) is added to the set of predictable tokens $\mathcal{A}$, and appended to the ground truth sequence: $\mb{y} \in \mathcal{A}^{L^y+1}$ with $\mathcal{A} = \mathcal{A}^c \cup \{ \text{<e>} \}$ and $\mb{y}_{L^y+1} = \text{<e>}$.

 Regarding the notations, given that matrices change over the decoding iterations, we opted for the following convention. The exponent specifies the version of the matrix at a given time, and the index specifies a vector from that matrix. For instance, in Figure~\ref{fig:dan-archi}, $\mb{q}^t$ is the matrix of query token vectors at decoding iteration $t$ made up of $t$ vectors indexed from $0$ to $t-1$. Given that the vector $\mb{q}_i^t$ at index $i$ is the same for all $t$, it can be referred to as $\mb{q}_i$, omitting the exponent for readability.
In the following, we describe the different stages implied by the DAN by referring to the circled numbers indicated in Figure~\ref{fig:dan-archi}.\\

\circled{1} The DAN encoder, noted $\text{Enc}_\text{DAN}$, extracts 2D features $\mb{f}=\text{Enc}_\text{DAN}(\mb{X})$ ($\mb{f} \in \mathbb{R}^{H_f \times W_f \times C_f}$) from the input image by reducing the height by 32 and the width by 8 $(H_f=\frac{H}{32}$, $W_f=\frac{W}{8}$ and $C_f=256)$. This encoder was first proposed in~\cite{Coquenet2023a} as a generic feature extractor for handwritten text recognition able to deal with images of variable sizes. All details can be found in the original paper~\cite{Coquenet2023a}.\\
\indent \circled{2} For transformer needs, positional information is injected into the image features $\mb{f}$ through additive 2D positional encoding $\mb{\pi}_\text{2D}$~\cite{Singh2021}. Height and width are each encoded over half of the vector channels. The result is reshaped to a one-dimensional representation: $\mb{g} = \text{flatten}(\mb{f} + \mb{\pi}_\text{2D})$.\\
\indent \circled{3} For each iteration $t$, the previously predicted tokens are put together to be used by the model to know which character token should be predicted next: $\hat{\mb{y}}^t = [\hat{\mb{y}}_0, \dots, \hat{\mb{y}}_{t-1}]$. A special start-of-transcription token (<s>) is used to initiate the prediction process: $\hat{\mb{y}}_0 = \text{<s>}$.\\
\indent \circled{4} A trainable weight matrix $\mb{E} \in \mathbb{R}^{(|\mathcal{A}|+1) \times C_f}$ is used to map each token of the vocabulary $\mathcal{A}$ to a vector of dimension $C_f$, leading to $\mb{e}^t \in \mathbb{R}^{t \times C_f}$, with $\mb{e}^t_i = \mb{E}_{\hat{\mb{y}}_i}$. One can note the additional embedding vector (+1), which corresponds to the start-of-transcription token (<s>).\\
\indent \circled{5} As the transformer attention mechanism is permutation equivariant, one-dimensional positional encoding $\mb{\pi}^t_\text{1D}$~\cite{Vaswani2017} is added to these embeddings to preserve the order information of the tokens, resulting in the query sequence $\mb{q}^t = \mb{e}^t + \mb{\pi}^t_\text{1D}$.\\
\indent \circled{6} The query tokens $\mb{q}^t$ are processed through the DAN decoder, noted $\text{Dec}_\text{DAN}$, which corresponds to 8 stacked transformer decoder layers preserving the number of channels $C_f$. The goal is to extract relevant information from the flattened image features $\mb{g}$, \ie{}, a latent representation $\mb{o}^t_{t-1}$ of the next character to predict: $\mb{o}^t = \text{Dec}_\text{DAN}(\mb{q}^t, \mb{g})$.\\
\indent \circled{7} Based on the last output of the decoder $\mb{o}_{t-1}$, a score vector $\mb{s}_t \in \mathbb{R}^{|\mathcal{A}|}$ is computed through a linear projection layer of trainable weights $\mb{W} \in \mathbb{R}^{|\mathcal{A}| \times C_f}$ to associate a score with each predictable token:
\begin{equation}
    \label{eq:score-dan}
   \mb{s}_t = \mb{W}\mb{o}_{t-1}.
\end{equation}

\indent \circled{8} The predicted token $\hat{\mb{y}}_t$ corresponds to the one with the highest score: $\hat{\mb{y}}_t = \argmax(\mb{s}_t)$.\\

Steps \circled{3} to \circled{8} are repeated until the specific end-of-transcription token (<e>) is predicted.

The DAN is trained using the cross-entropy loss $\mathcal{L}_\text{CE}$ over the sequence of logits of the scores obtained through softmax activation:

\begin{equation}
    \label{eq:loss-dan}
    \mathcal{L}_\text{DAN} = \sum_{t=1}^{L^y+1} \mathcal{L}_\text{CE}(\mb{y}_t,\mb{p}_t),
\end{equation}
where $ \displaystyle \mb{p}_{t,i} = \frac{e^{\mb{s}_{t,i}}}{\sum_j e^{\mb{s}_{t,j}}}$ is the predicted probability associated with token $i$ at time $t$.

\subsection{Faster DAN}

The Faster DAN~\cite{Coquenet2023c} was proposed as a decoding alternative to speed up the inference stage. 
It aims to parallelize the text line predictions by relying on a two-stage approach. First, an autoregressive process predicts the first character of each line, following the original DAN strategy. 
Then, all lines are completed in parallel, reducing the total number of iterations. The main difference appears in that second decoding stage. 
Instead of using the output corresponding to the last token $\mb{o}_{t-1}$ only, the token prediction is performed using the outputs corresponding to the last token of each text line. 
In this way, for $N$ text lines, there are $N$ token predictions per iteration. Equation~\ref{eq:score-dan} becomes: $\displaystyle \mb{s}^l_t = \mb{W}\mb{o}^l_{t-1}$,
where $\mb{s}^l_t$ is the score vector associated with the token prediction of the next character for line $l$ at iteration $t$, and $\mb{o}^l_{t-1}$ is the transformer output corresponding to the last predicted token for line $l$.

Although the Faster DAN showed interesting inference speed improvements, it results in a loss of past context when predicting a character. 
Indeed, instead of benefiting from the whole past context as is the case for the DAN, the Faster DAN has a partial view of the past and future due to the parallel line predictions, which reduces its ability to model the language.

The strategy we propose aims to get the best of both worlds: improving context modeling by processing successive queries altogether through multi-token predictions, while reducing prediction time through parallelization.
\section{\metadan{}}

\begin{figure}[ht!]
    \centering
    \begin{subfigure}[t]{0.22\textwidth}
        \centering
        \includestandalone[width=\textwidth]{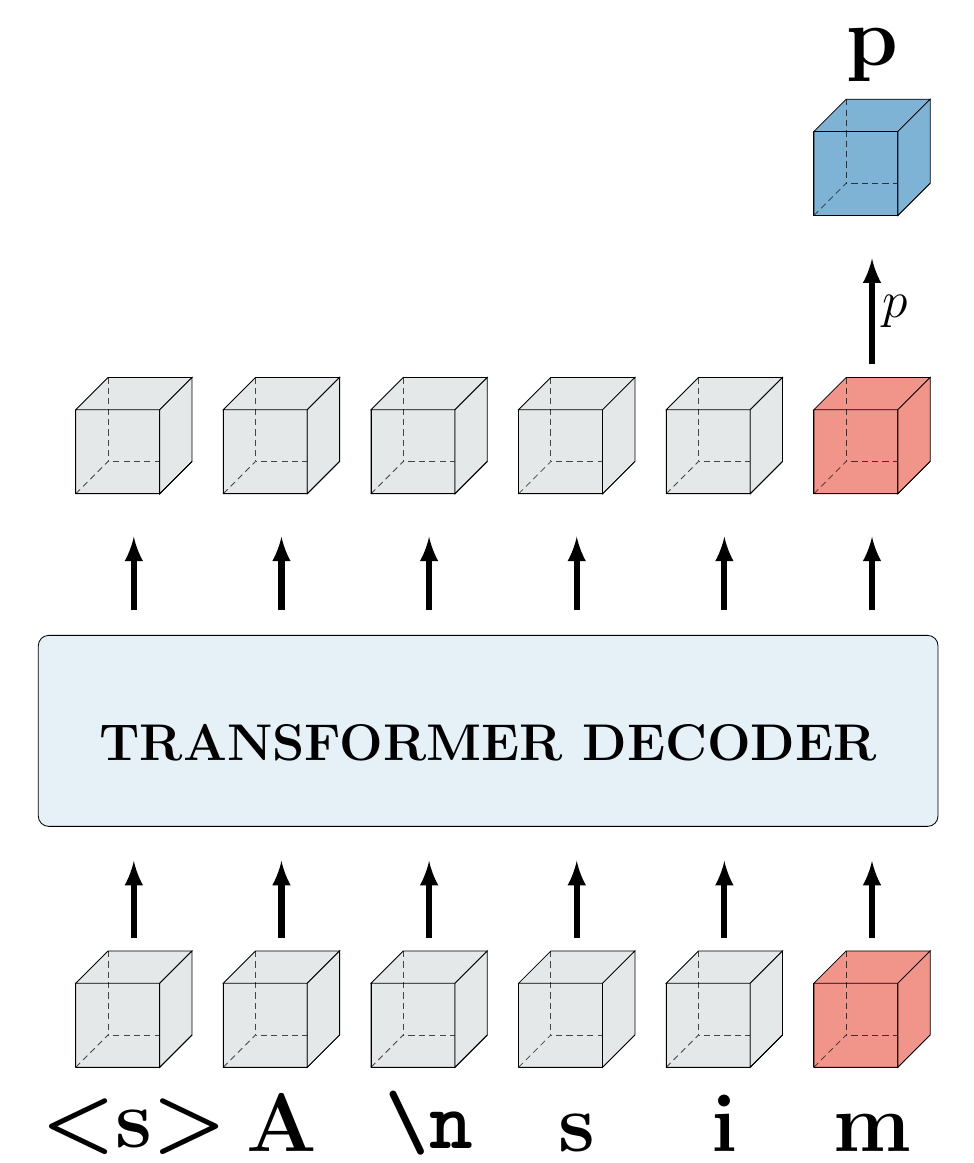}
        \caption{DAN \cite{Coquenet2023b}}
        \label{fig:comparison-prediction-dan}
    \end{subfigure}
    \hfill
    \begin{subfigure}[t]{0.25\textwidth}
        \centering
        \includestandalone[width=\textwidth]{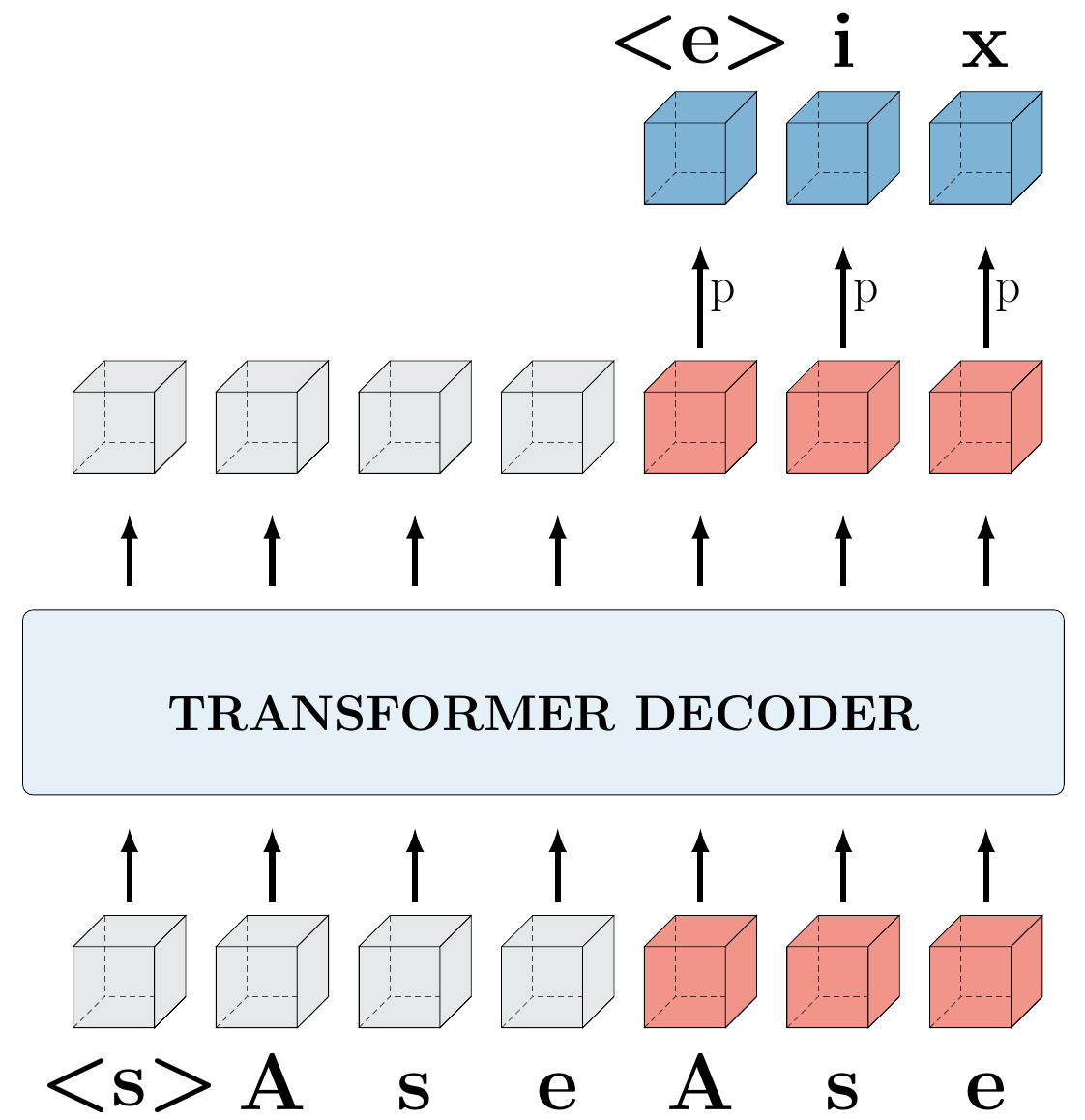}
        \caption{Faster DAN \cite{Coquenet2023c}}
        \label{fig:comparison-prediction-faster-dan}
    \end{subfigure}
    \hfill
    \begin{subfigure}[t]{0.22\textwidth}
        \centering
        \includestandalone[width=\textwidth]{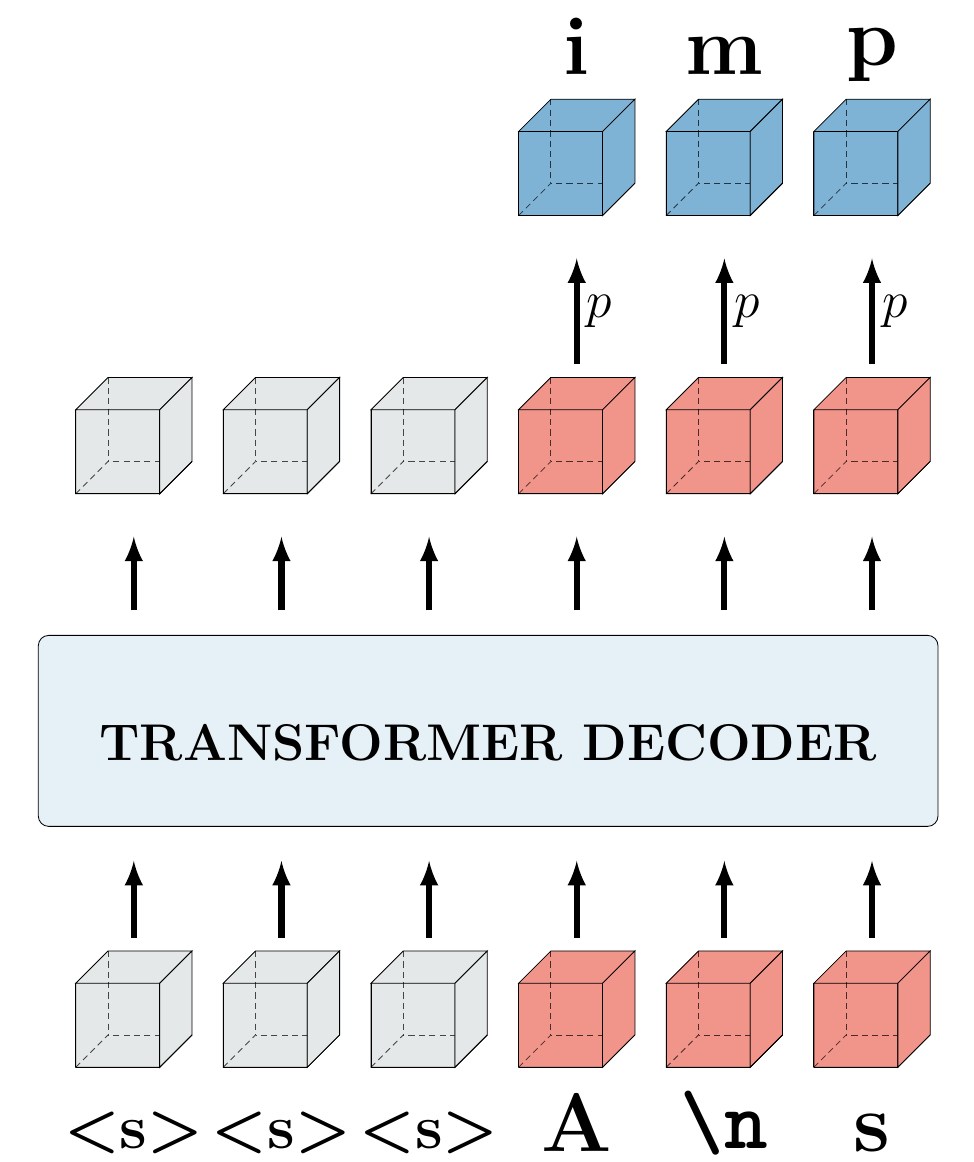}
        \caption{W-DAN ($w=3$)}
        \label{fig:comparison-prediction-w-dan}
    \end{subfigure}
    
    \vspace{0.2cm}
    
    \begin{subfigure}[t]{0.23\textwidth}
        \centering
        \includestandalone[width=\textwidth]{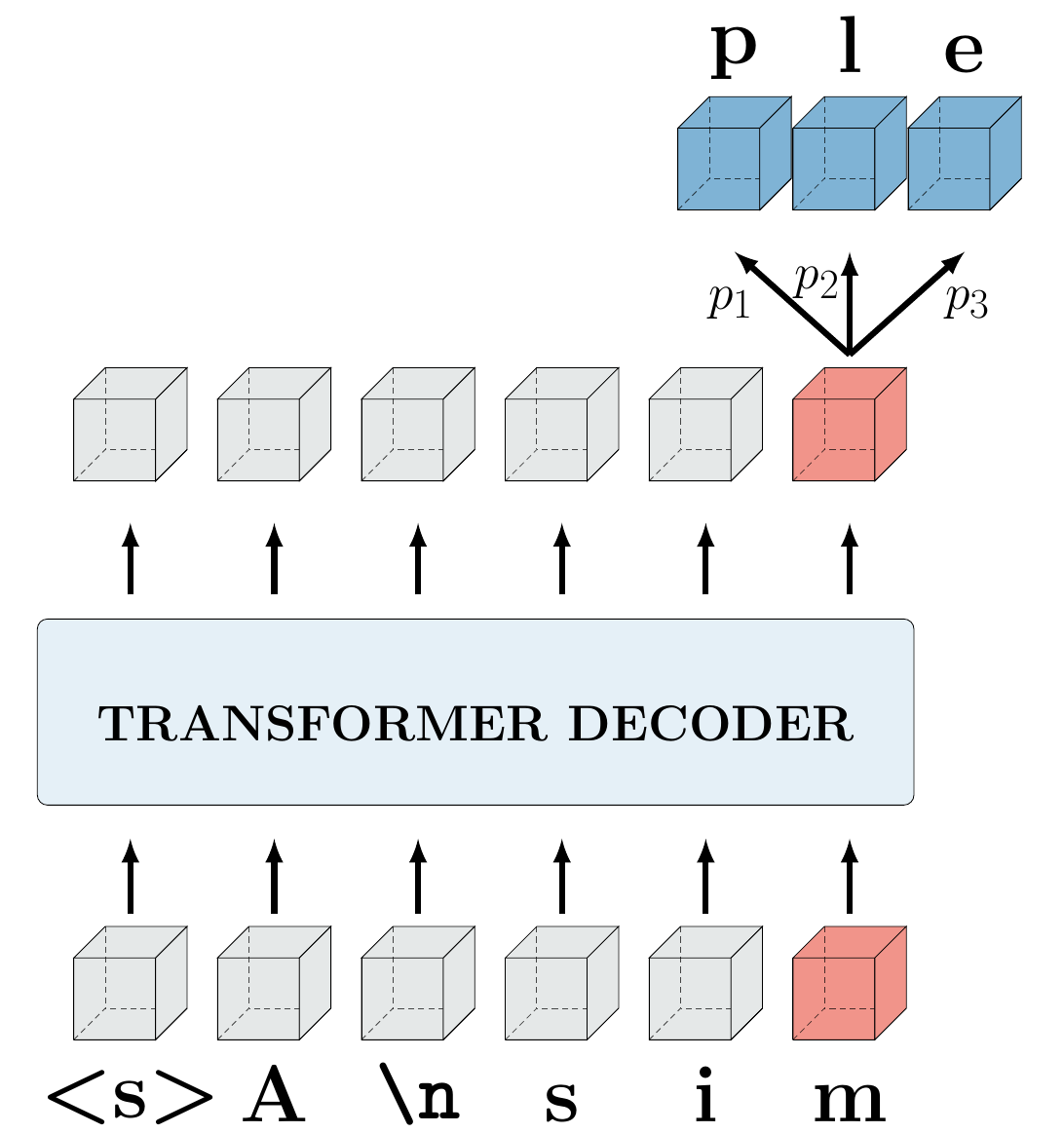}
        \caption{MT-DAN ($m=3$)}
        \label{fig:comparison-prediction-mt-dan}
    \end{subfigure}
    \hspace{2cm}
    \begin{subfigure}[t]{0.32\textwidth}
        \centering
        \includestandalone[width=0.85\textwidth]{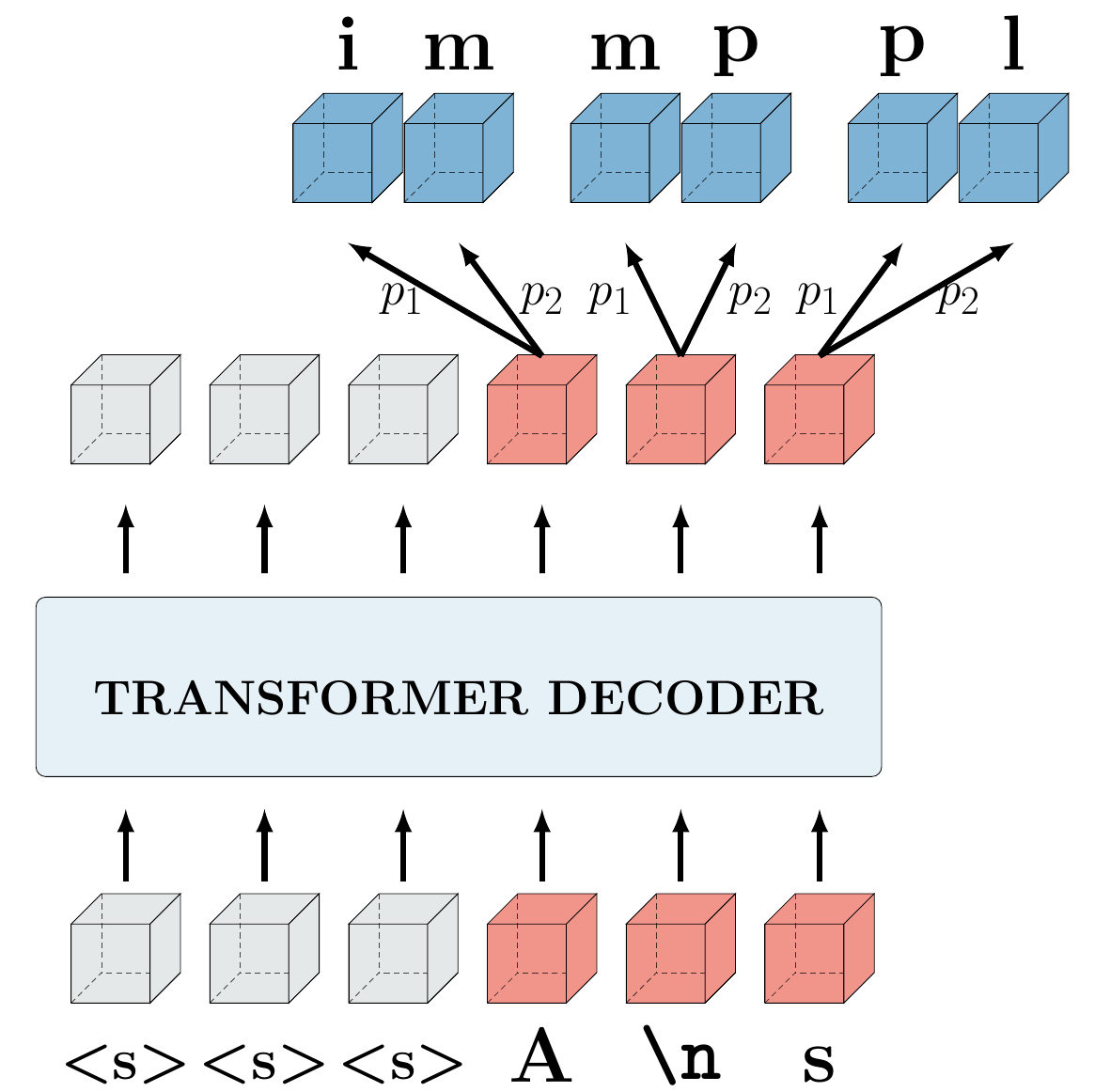}
        \caption{Meta-DAN ($w=3$, $m=2$)}
    \label{fig:comparison-prediction-metadan}
    \end{subfigure}
    \caption{Comparison of the prediction strategies for the target sequence "A\escape{n}simple\escape{n}example.", under three aspects: 1) the number of predictions per query (single for DAN, Faster DAN and \wdans{}); 2) the number of queries processed at once per decoding iteration (single for DAN and \mtdans{}); 3) the available context (start of each line for Faster DAN; full past for the others). 
    The \metadans{} groups together the advantages of each of these points: it processes multiple queries at once, each one leading to several token predictions, and benefits from the full past context.
    }
    \label{fig:comparison-prediction}
\end{figure}

We propose two new approaches for the autoregressive prediction of text tokens in the context of end-to-end handwritten text recognition at page level. 
For both, the idea is to predict multiple tokens at each decoding iteration to reduce the number of decoding steps, while preserving the whole past context. We assume this is a key element to reduce the number of recognition errors through language modeling. 
First, we propose to perform multiple predictions from the same query $\mb{q}_{t-1}$, to predict the $m$ next tokens at once. This is referred to as the \mtdan{} (\mtdans{}).
Second, we propose a query approach based on a window mechanism to process multiple queries at once, each associated with a single prediction. This is referred to as the \wdan{} (\wdans{}).
Finally, we design a global strategy including these two approaches, called \metadan{} (\metadans{}).
A visual comparison of the prediction processes is depicted in Figure~\ref{fig:comparison-prediction}. 

\subsection{\mtdans{}}
Inspired by recent works on multi-token predictions for Natural Language Processing (NLP)~\cite{Gloeckle2024}, we propose the \mtdan{} (\mtdans{}). It relies on the following process. Based on the last query output $\mb{o}_{t-1}$, $m$ projection heads are used to compute the scores for the $m$ next tokens, with $m$ being a new hyperparameter.
Equation~\ref{eq:score-dan} becomes:

\begin{equation}
    \mb{s}^{(\mb{t-1})}_{t+k} = \mb{W}^k\mb{o}_{t-1},
\end{equation}
where $\mb{s}^{(\mb{t-1})}_{t+k}$ is the score vector associated with the token prediction of the $k^\text{th}$ next token computed at iteration $t$. The exponent $(\mb{t-1})$ indicates that this prediction was performed based on query indexed by $t-1$. $\mb{W}^k$ is the weight matrix corresponding to the $k^\text{th}$ projection head, and $\mb{o}_{t-1}$ is the transformer output corresponding to the last query token $\mb{q}_{t-1}$. In this way, $m$ predictions are performed per iteration following:

\begin{equation}
    \hat{\mb{y}}^{(\mb{t})}_{t+k} = \argmax(\mb{s}^{(\mb{t})}_{t+k}),
\end{equation}
with $k \in [0, \dots, m-1]$.

As one can note, several predictions $\hat{\mb{y}}^{(\mb{t})}_{t+k}$ are associated with the same target token $\mb{y}_{t+k}$. This is not an issue during training; we train the \mtdans{} using the cross-entropy loss over the whole token sequence, as for the original DAN. It involves adding $m-1$ extra <e> tokens in order for all text tokens to have a target for each of the projection heads, leading to the following loss:

\begin{equation}
    \mathcal{L}_\text{\mtdans{}} = \sum_{t=1}^{L^y+1} \sum_{k=0}^{m-1} \mathcal{L}_\text{CE}(\mb{y}_{t+k},\mb{p}^{(\mb{t})}_{t+k}),
\end{equation}
with $\mb{p}^{(\mb{t})}_{t+k} = \softmax(\mb{s}^{(\mb{t})}_{t+k})$.

Predicting multiple tokens $\hat{\mb{y}}_{t+k}$ from a unique output $\mb{o}_{t-1}$ can theoretically lead to saturation as $m$ increases. Indeed, the fixed-size vector $\mb{o}_{t-1}$ must represent the next $m$ text tokens at once. 
To solve this issue, we propose another way to predict multiple tokens per iteration, which consists in processing several consecutive queries at once.

\subsection{\wdans{}}

To solve the \mtdans{} output compression issue, we propose the \wdan{} (\wdans{}). Instead of relying on a single output to predict $m$ tokens, we opt for using the output of the last $w$ predictions to predict the $w$ next tokens, $w$ being a new hyperparameter. In this way, it solves the compression issue as there are as many outputs as there are predictions. The decoding process of the \wdans{} acts as a sliding window over the query tokens, processing them by chunks of $w$ tokens at once, leading to $w$ new outputs.
More precisely, each output $\mb{o}_{t-q}$ is responsible for the prediction of a single token $\hat{\mb{y}}_{t-q+w}$, with $q \in [1, \dots, w]$.
Equation~\ref{eq:score-dan} becomes:

\begin{equation}
    \mb{s}_{t-q+w} = \mb{W}\mb{o}_{t-q},
\end{equation}
with
\begin{equation}
    \hat{\mb{y}}_{t-q+w} = \argmax(\mb{s}_{t-q+w}).
\end{equation}

It has to be noted that all scores $\mb{s}_{t-q+w}$ are computed using the same projection head of weights $\mb{W}$ as they are applied independently on each $\mb{o}_{t-q}$. 
In addition, it is necessary to initialize the decoding process with $w$ initial tokens to predict the first $w$ tokens. 
We arbitrarily choose to use the same start-of-transcription token $w$ times ($\hat{\mb{y}}_0 = \cdots = \hat{\mb{y}}_{w-1} = \text{<s>}$). 
We also added $n_\text{e}$ extra end-of-transcription tokens at the end of the target sequence so as to make the target length divisible by $w$. It leads to the following loss:

\begin{equation}
    \mathcal{L}_\text{\wdans{}} = \sum_{t=w}^{L^y+1+n_\text{e}} \mathcal{L}_\text{CE}(\mb{y}_{t},\mb{p}_{t}).
\end{equation}

The major advantage of both presented approaches compared to the DAN and the Faster DAN is that the $k$ or $w$ predictions are consecutive and performed altogether: the context is enlarged when the prediction takes place. The model must focus on a broader context to predict multiple successive tokens. For instance, for $k=3$ (or $w=3$), the model focuses on predicting $\{ \hat{\mb{y}}_t, \hat{\mb{y}}_{t+1}, \hat{\mb{y}}_{t+2}\}$ at once. This way, the prediction of $\hat{\mb{y}}_t$ is impacted by the future context $\{\hat{\mb{y}}_{t+1}, \hat{\mb{y}}_{t+2}\}$, which improves the ability to model the language at prediction time compared to DAN.

\subsection{\metadans{}}

We propose to take advantage of both presented approaches by combining them: this is referred to as the \metadan{} (\metadans{}). Thus, the \metadans{} is parametrized by two hyperparameters $m$ and $w$, indicating the number of projection heads and the window size, respectively. In other words, at each iteration $t$, a window of $w$ query tokens $\{\mb{q}_{t-w}, \dots, \mb{q}_{t-1}\}$ are processed at once, each one leading to $m$ predictions. More precisely, the output of the query token $\mb{q}_{t-q}$, \ie{} $\mb{o}_{t-q}$, is responsible for the prediction of $\{ \hat{\mb{y}}_{t-q+w}, \dots, \hat{\mb{y}}_{t-q+w+m-1} \}$ as follows:

\begin{equation}
    \hat{\mb{y}}^{(\mb{t-q})}_{t-q+w+k} = \argmax(\mb{s}^{(\mb{t-q})}_{t-q+w+k}),
\end{equation}
with
\begin{equation}
    \mb{s}^{(\mb{t-q})}_{t-q+w+k} = \mb{W}^k\mb{o}_{t-q},
\end{equation}
where $\mb{s}^{(\mb{t-q})}_{t-q+w+k}$ is the score associated with token $\hat{\mb{y}}_{t-q+w+k}$ by the $k^\text{th}$ projection head of the $q^\text{th}$ query token in the query window, and with $k \in [0, \dots, m-1]$ and $q \in [1, \dots, w]$. 
The  \metadans{} involves initializing the decoding process with $w$ <s> tokens, as for the \wdans{}, and adding $n_e$ extra <e> so that the last query has a target prediction for all its associated projection heads. Regarding training, equation~\ref{eq:loss-dan} becomes:
\begin{equation}
    \mathcal{L}_\text{\metadans{}} = \sum_{t=w}^{L^y+1+n_\text{e}}  \sum_{k=0}^{m-1} \mathcal{L}_\text{CE}(\mb{y}_{t+k}, \mb{p}^{(\mb{t})}_{t+k})
\end{equation}

As one can note, several query tokens can be in charge of the same prediction. 
This is shown in Figure~\ref{fig:comparison-prediction-metadan} with an example with $w=3$ and $m=2$: the "m" and "p" are predicted twice, from two different queries. 
In this work, we limit ourselves to using the prediction of the first projection head for the $w-1$ first queries, and the prediction of all projection heads for the last one.
Although discarded, we assume that training to predict these extra characters encourages the network to improve language modeling. 

In summary, the \metadans{} is a general formulation for which one can distinguish three specific cases depending on the hyperparameters: 
\begin{itemize}
    \item Choosing $m=1$ and $w=1$ consists in the DAN approach.
    \item Choosing $m>1$ and $w=1$ results in the \mtdans{} approach.
    \item Choosing $m=1$ and $w>1$ leads to the \wdans{} approach.
\end{itemize}

\subsection{Impact on the computational complexity}

\begin{figure*}[ht!]
    \captionsetup[subfigure]{justification=centering}
    \begin{subfigure}[t]{0.3\textwidth}
        \includestandalone[width=\textwidth]{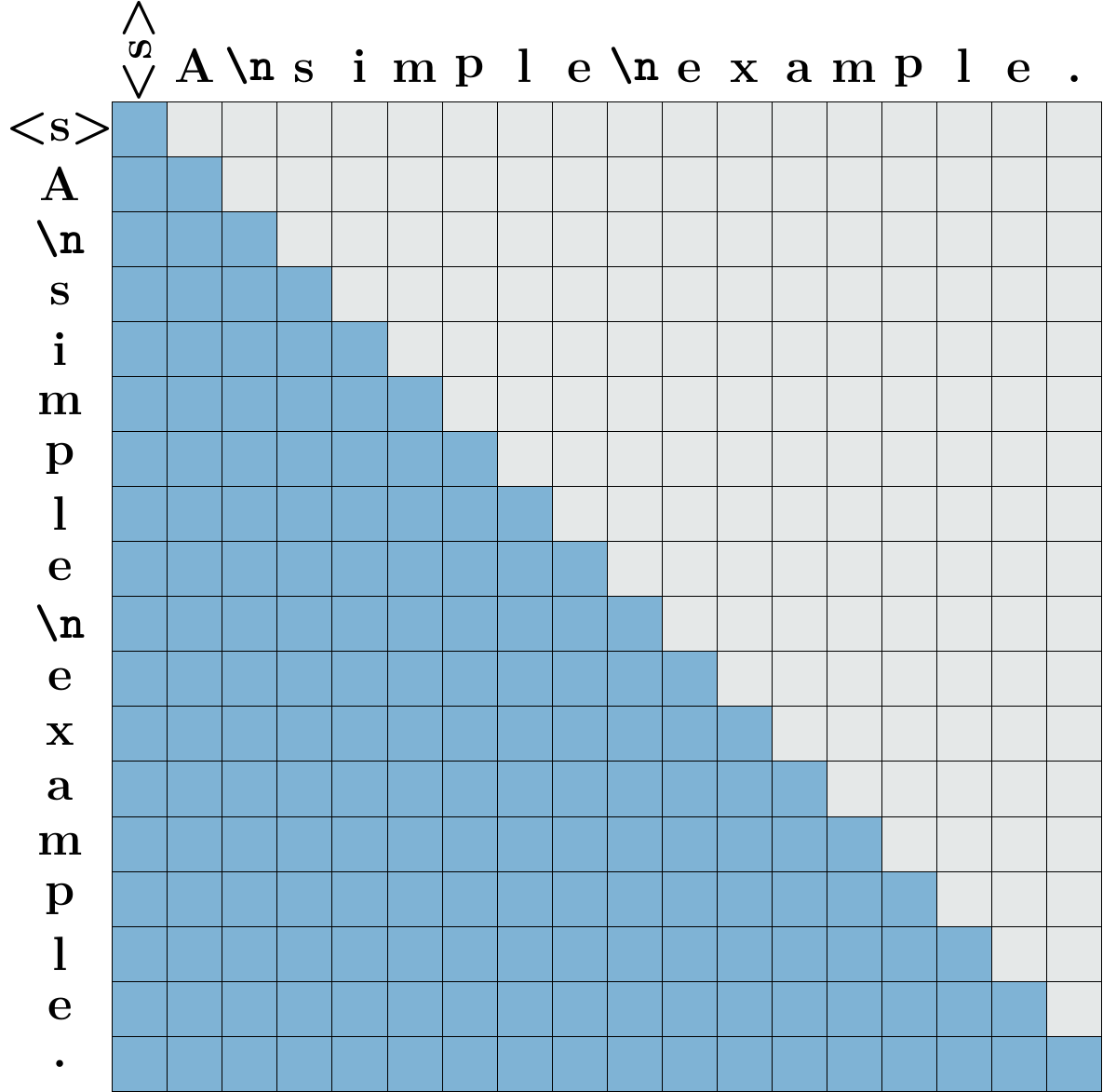}
        \caption{DAN \cite{Coquenet2023b}/\mtdans{} (ours)}
    \end{subfigure}
    \hfill
    \begin{subfigure}[t]{0.3\textwidth}
        \includestandalone[width=\textwidth]{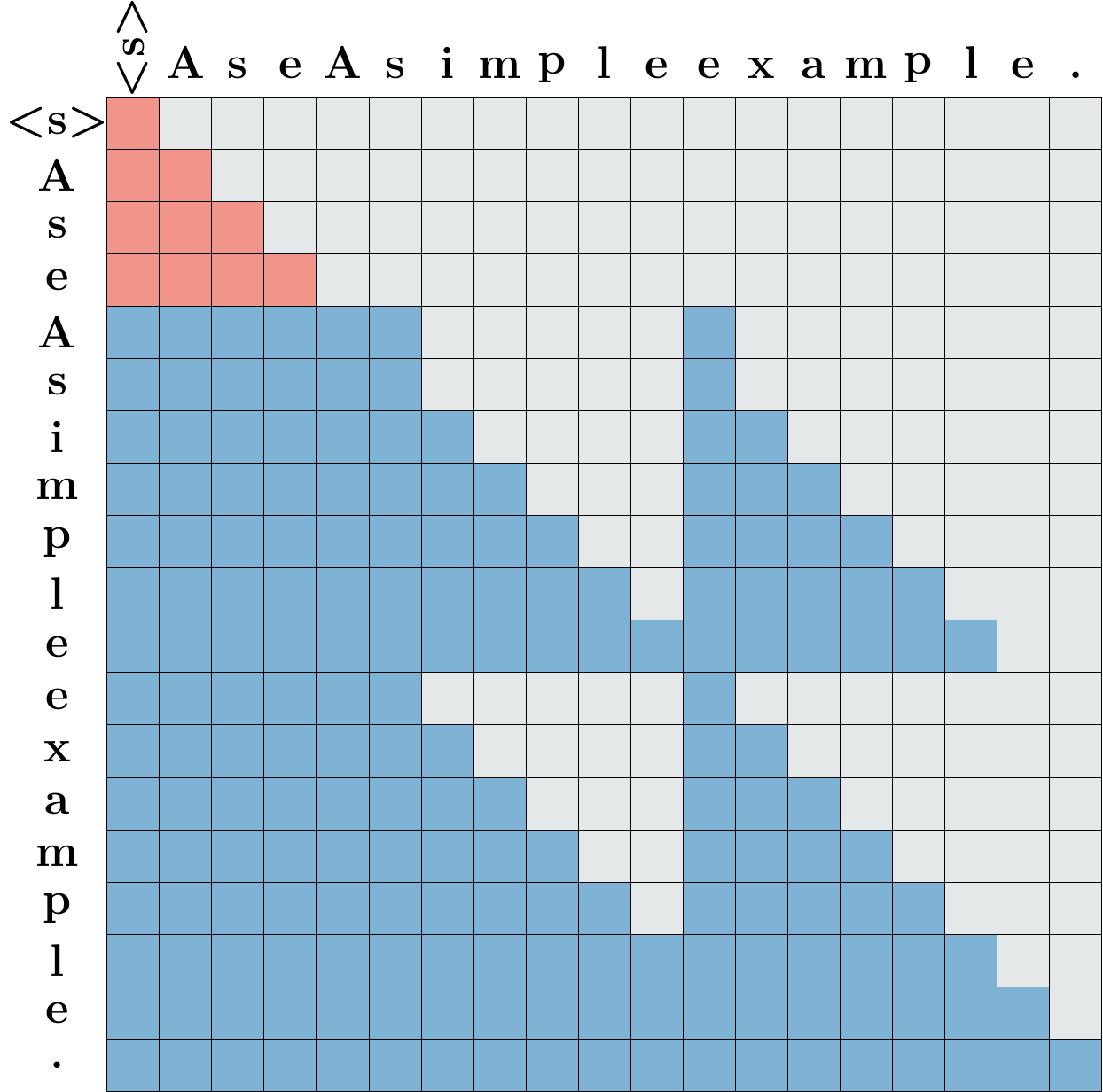}
        \caption{Faster DAN \cite{Coquenet2023c}}
    \end{subfigure}
    \hfill
    \begin{subfigure}[t]{0.3\textwidth}
        \includestandalone[width=\textwidth]{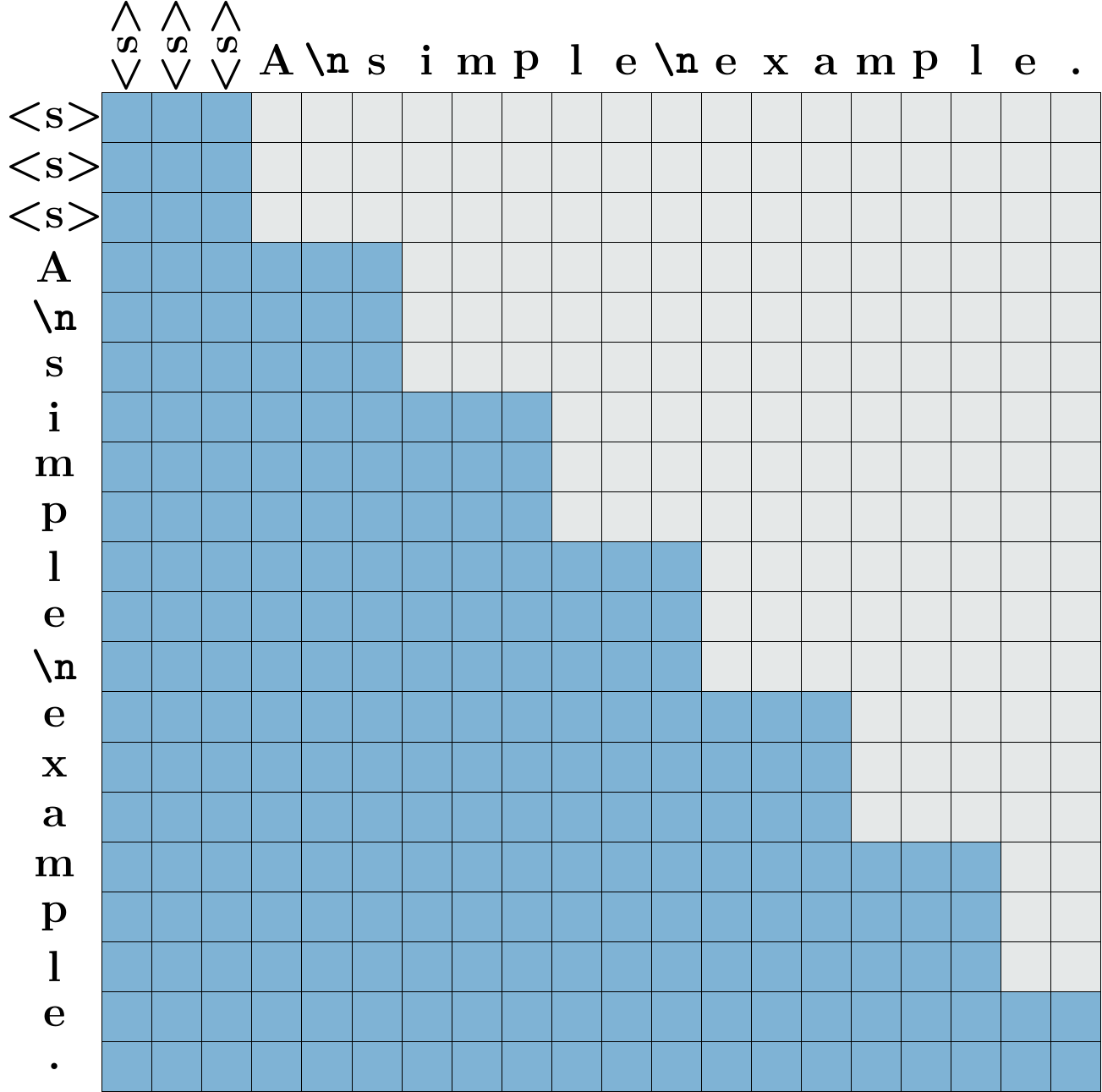}
        \caption{\wdans{} / \metadans{} \\(ours, for $w=3$)}
    \end{subfigure}
    \caption{Comparison of attention context for the target sequence "A\escape{n}simple\escape{n}example.". The blue color indicates which tokens can be attended (column) when processing a given token (line). Grey cells corresponds to ignored tokens (\ie, not yet predicted tokens). Red color highlights the first pass of the Faster DAN.}
    \label{fig:comparison-attention}
\end{figure*}

To further give insights into the main differences between all these prediction strategies, Figure~\ref{fig:comparison-attention} provide visuals of the attending maps (transformer masks), stating which token can access to which other tokens for the masked self-attention.

Regarding the \mtdans{} approach, the attending maps of the DAN and the \mtdans{} are identical, no matter the value of $m$; this means that they both imply the same computations in the transformer decoder layers. 
In addition, contrary to the NLP application, which relies on the Byte-Pair Encoding algorithm leading to large vocabularies, the addition of extra projection heads does not significantly impact the global computational complexity as we are working at character level, limiting the dimensions of the projection weight matrices.

Concerning the \wdans{} approach (and thus the \metadans{}), the attending maps show that there are more interactions between the tokens in the self-attention layers, resulting in more computations, theoretically. 
However, to our knowledge, the current implementations of the transformer rely on masks to "hide" the forbidden tokens (usually the future) \textit{after} the query-key multiplication takes place. 
In this way, this does not impact the computation time in practice since computations are performed over all token pairs, whether they are really used or not.

No matter the approach, the prediction process can be parallelized at training time using teacher forcing~\cite{TeacherForcing}, \ie{}, by injecting the whole ground truth instead of the autoregressive previous predictions, so that all decoding iterations are performed at once. In this way, there is no extra benefit for the training stage. 
The major impact is positive and related to the time complexity at prediction time when parallelizing the computation of token predictions.

\section{Experimental conditions}

\subsection{Datasets}

We select 10 public datasets with high diversity to make our evaluation robust: IAM~\cite{IAM}, READ-2016~\cite{READ2016}, Esposalles~\cite{ESPOSALLES}, BRESSAY~\cite{BRESSAY}, ScribbleLens~\cite{SCRIBBLELENS}, Esparchos~\cite{EPARCHOS}, CASIA-2~\cite{CASIA}, MAURDOR~\cite{MAURDOR}, RIMES-2009~\cite{RIMES2009} and RIMES-2024~\cite{RIMES2024}. 
As detailed in Table~\ref{tab:dataset-page}, they cover 8 languages, three alphabets (Latin, Greek, and Chinese), and 6 centuries. They also involve many writing configurations (from a single writer to more than a thousand), various sizes of character sets, as well as a variable amount of training data.

\begin{table}[ht!]
    \centering
    \caption{Detailed features of evaluated datasets (page-level).}
    \resizebox{\textwidth}{!}{
    \begin{tabular}{|c|c|c|c|c|c|c|c|c|}
    \hline
     \multirow{2}{*}{Dataset} & \multirow{2}{*}{Train} & \multirow{2}{*}{Val} & \multirow{2}{*}{Test} & Charset & Average &\multirow{2}{*}{Language} & \multirow{2}{*}{Century} & \multirow{2}{*}{Writers} \\
    & & & & size & \# char/page & & &\\
    \hline        
        IAM (paragraph)~\cite{IAM} & 747 & 161 & 336 & 80 & 371 & English & 20\textsuperscript{th}-21\textsuperscript{st} & 657\\
        RIMES-2009~\cite{RIMES2009} & 1,050 & 100 & 100 & 108 & 588 & French & 21\textsuperscript{st} & $\approx 400$\\
        RIMES-2024~\cite{RIMES2024}  & 4,500 & 505 & 600 & 123 & 567 & French & 21\textsuperscript{st} & $\approx$ 1,400\\
        READ-2016~\cite{READ2016} & 350 & 50 & 50 & 89 & 468 & German & 15\textsuperscript{th}-19\textsuperscript{th} & unknown\\
        Esposalles~\cite{ESPOSALLES}  & 75 & 25 & 25 & 62 & 1,748 & Catalan &  17\textsuperscript{th} & 1\\
        BRESSAY~\cite{BRESSAY} & 647 & 154 & 199 & 109 & 2,612 & Portuguese & 21\textsuperscript{st} & 1,000\\ 
        ScribbleLens~\cite{SCRIBBLELENS} & 164 & 14 & 21 & 85 & 1,410 & Dutch & 16\textsuperscript{th}-18\textsuperscript{th} & 85\\
        Eparchos~\cite{EPARCHOS} & 80 & 20 & 20 & 157 & 962 & Greek & 16\textsuperscript{th} & 2\\
        CASIA-2~\cite{CASIA} & 3,600 & 475 & 1,015 & 2,704 & 273 & Chinese & 21\textsuperscript{st} & 1,019\\
        MAURDOR-C3~\cite{MAURDOR} & 1,006 & 148 & 166 & 134 & 491 & French/English & 21\textsuperscript{st} & > 130\\
        MAURDOR-C4~\cite{MAURDOR} & 721 & 111 & 114 & 127 & 675 &  French/English & 21\textsuperscript{st} & > 90\\
    \hline
    \end{tabular}
    }
    \label{tab:dataset-page}
\end{table}

We used the standard split into training, validation, and test sets when existing. 
To our knowledge, we are the first to use the recently released RIMES-2024 dataset, so we propose a split, randomly sampling for training, validation and test by respecting the split provided at line level in 2011~\cite{RIMES2011}. 
We discarded images from the "Q" category, which corresponds to forms, since they do not provide the whole transcript. 
For ScribbleLens, we removed 10\% of the training samples to create a validation set by randomly picking from each of its data sources. 
For Eparchos, we propose a split at page level, by taking 80\% of the samples for training, 10\% for validation, and 10\% for test. 
\ref{sec-appendix-dataset} provides the splits at line level, used for pre-training purposes, for the datasets providing line-level annotations. This section also includes document image examples.

\subsection{Training strategy}
If not stated otherwise, we follow the main guidelines described in~\cite{Coquenet2023b}:
\begin{itemize}
    \item The encoder is pre-trained using a CTC-based line-level training strategy. An adaptive max-pooling layer is added on top of the encoder to collapse the vertical axis, leading to a one-dimensional sequence of features. Then, a classification layer is applied to generate a probability lattice. If line-level annotations are provided in the dataset, we use a ratio of 25\% real / 75\% synthetic images (100\% synthetic otherwise).
    \item Data augmentation is applied with a probability of 90\%, including resolution change, color jittering, Gaussian noising, Gaussian blurring, elastic distortion, dilation, erosion, and perspective transformation. 
    \item The page-level training relies on a curriculum synthetic document generation strategy. The generated images contain more and more text lines through training. The ratio of synthetic documents slowly decreases from 90\% to 10\%. Following~\cite{Constum2024}, we rely on the 1001-fonts.com website to get a large collection of fonts from which we randomly sample when rendering a synthetic text line. The text used for synthetic generation is from the training set of the target dataset.
    \item Line-level pre-training is carried out using 5M training samples; page-level training (including curriculum) is performed over 1M samples. Training and pre-training are performed using the Adam optimizer with an initial learning rate of $10^{-4}$, and automatic mixed-precision. 
    \item Preprocessing: images are scaled to reach 150 dpi resolution. No post-processing: we do not use any external data or language model.
\end{itemize}

\subsection{Evaluation}
We compute the standard Character Error Rate (CER) and Word Error Rate (WER) to evaluate the text recognition performance. Both metrics are computed using the Levenshtein distance~\cite{Levenshtein}, which counts the number of deletion, addition, and substitution operations required to go from the prediction sequence to the ground truth sequence, normalized by the length of the ground truth sequence. While the CER considers the prediction and the ground truth as sequences of characters, the WER considers them as sequences of words.
All results are given for the test sets.
It has to be noted that page-level approaches are sensitive to reading order; it means that correct text recognized in a wrong order will be considered as incorrect.
For each model, we compute the average prediction time for one test image when using a single A100 GPU and a mini-batch size of 1.
We set the maximum number of decoding iterations for all datasets to 5,000 to cover the requirements of all evaluated datasets. This way, we do not rely on dataset-specific properties, which is closer to a real-case scenario. 
It has to be noted that this choice results in recognition metrics that can be highly impacted by the prediction loops that can occur, \ie{}, when the model predicts again and again the same text line.

\section{Experiments}

In this section, we evaluate the proposed approaches (\wdans{}, \mtdans{}, and \metadans{}) on the IAM and BRESSAY datasets to select the best approach and hyperparameters. 
Then, we apply them to the other 9 datasets to validate the robustness of the approach. 
While relying on the original DAN architecture~\cite{Coquenet2023b}, we propose an enlarged version of the encoder, which we refer to as FCN-M as it shows better recognition performance (details in~\ref{sec:appendix-encoder}).

\subsection{\wdans{}}

Table~\ref{tab:mtw-dan-results}(a) provides the results of the \wdans{} approach for several window sizes $w$. 
It highlights that it is possible to predict up to 10 character tokens at once, while being competitive in terms of CER. 
It is especially true for the BRESSAY dataset for which $w=10$ leads to a small increase of only 0.23 points of CER. 
However, results degrade significantly for $w=25$ with a CER of 6.56\% instead of 4.23\% for IAM, and 4.37\% instead of 2.43\% for BRESSAY. 

As one can note, it is possible to reduce the average prediction time per sample nearly linearly with respect to the window size. 
Regarding BRESSAY prediction times, it appears that the new implementation has impacted the prediction time for large text predictions, leading to a lag in the speed factor improvement.
A main observation is that the model is harder to train as $w$ increases. We assume that this is due to the inherent complexity of the self-attention layer, which grows with the length of the query sequence. Indeed, the more query tokens are processed altogether, the greater the required effort to coordinate and focus on the right parts of the image features.
In this way, we needed to increase the amount of training data to reach convergence for $w=10$ and $w=25$ up to 2M and 3M, respectively.

\begin{table*}[ht!]
    \centering    
    \caption{Comparison of \wdans{} and \mtdans{} performance with respect to the window size $w$ and to the number of token heads $m$, respectively, on the IAM and BRESSAY test sets.}
    \label{tab:mtw-dan-results}

    \begin{threeparttable}[b] 

        \begin{minipage}[c]{0.48\linewidth} 
            \centering
            {\scriptsize (a) \wdans{} results.}\\
            \vspace{0.05cm}
            \label{tab:w-dan-results}
            \resizebox{\linewidth}{!}{
            \begin{tabular}{|c|c|c|c|c|}
            \hline 
            \multirow{2}{*}{$w$}& \multicolumn{2}{|c|}{IAM} & \multicolumn{2}{|c|}{BRESSAY} \\
            \cline{2-5} 
            & CER (\%) & Time (s) &  CER (\%) & Time (s)\\
            \hline
            1 (DAN) & \textbf{4.23} & 3.39 & \textbf{2.43}  & 24.65 \\
            \hline
            2 & 4.67 & 1.94 & 2.91 & 20.86\\
            3 & 4.83 & 1.26 & 3.05 & 12.25\\
            4 & 4.75 & 0.97 & 2.63 & 8.63 \\
            5 & 4.87 & 0.77 & 3.02 & 6.63 \\
            10\tnote{*} & 5.07 & 0.40 & 2.66  & 2.95 \\
            25\tnote{*} & 6.56 & \textbf{0.18}  & 4.37 & \textbf{1.13}\\
            \hline
            \end{tabular}
            }
        \end{minipage}
        \hfill
        \begin{minipage}[c]{0.48\linewidth}
            \centering
            {\scriptsize (b) \mtdans{} results.}\\
            \vspace{0.05cm}
            \label{tab:mt-dan-results}
            \resizebox{\linewidth}{!}{
            \begin{tabular}{|c|c|c|c|c|}
            \hline
            \multirow{2}{*}{$m$}& \multicolumn{2}{|c|}{IAM} & \multicolumn{2}{|c|}{BRESSAY} \\
            \cline{2-5} 
            & CER (\%) & Time (s) &  CER (\%) & Time (s)\\
            \hline
            1 (DAN) & \textbf{4.23} & 3.39 & 2.43  & 24.65 \\
            \hline
            2 & 4.70 &  1.84 & 2.47 & 13.02 \\
            3 & 4.74 & 1.23  & 2.46 & 8.85 \\
            4 & 4.68 & 0.93  & 2.70 & 6.64 \\
            5 & 5.14 & 0.75  & 2.45 & 5.28\\
            10\tnote{*} & 5.06 & 0.39& \textbf{2.37} & 2.64 \\
            25\tnote{*} & 7.19 & \textbf{0.16} & 3.15 & \textbf{1.11}\\
            \hline
            \end{tabular}
            }
        \end{minipage}

        \begin{tablenotes}
            \scriptsize
            \item [*] More training data was used for $w=10$, $m=10$, $w=25$, and $m=25$
        \end{tablenotes}
    \end{threeparttable}
    
\end{table*}

\subsection{\mtdans{}}

For fair comparison, we evaluate the \mtdans{} approach for the same values of token heads $m \in \{1, 2, 3, 4, 5, 10, 25 \}$, as both $m$ for \mtdans{} and $w$ for \wdans{} refer to the number of predicted tokens per decoding iteration. Here, $m=1$ corresponds to the original DAN. 
The obtained results shown in Table~\ref{tab:mtw-dan-results}(b) are quite similar to those of the \wdans{}, with comparable recognition performance and prediction time. Neither approach seems to stand out from the other. 
Even for the extreme case of $m=25$, it appears that the \mtdans{} performs better for the BRESSAY dataset with a CER of 3.15\% compared to 4.37\% for the \wdans{} (with $w=25$), but it is the contrary for the IAM dataset: 7.19\% of CER compared to 6.56\% for the \wdans{}.
As for the \wdans{}, it required more training data for $m=10$ and $m=25$ to reach convergence. However, this is not for the same reason. We assume that, for the \mtdans{}, this is due to the compression task that is harder as $m$ increases. Indeed, $m$ character representations must be compressed into a single fixed-size vector. 
We also relied on a curriculum strategy at character level before the line-level curriculum strategy. 
We did not use it for the \wdans{} as it was not effective; we assume this is due to the way the tokens interact to make their predictions in this approach.

\paragraph{Improving the prediction policy}

\label{sec:mt-dan-dec-strat}
To enhance the capacity of the \mtdans{} approach, we propose and evaluate two variants, we called prediction policies, at inference time:

\noindent 1) The \textit{static} prediction policy consists in always predicting the $k$ next tokens at each decoding iteration 
($k \leq m$). 
Previous results, shown in Table~\ref{tab:mtw-dan-results}(b), relied on this naïve prediction policy, setting $k=m$.\\
2) The \textit{dynamic} prediction policy involves predicting between $1$ and $m$ tokens at once, based on a confidence threshold $\tau$. The number of predicted tokens must vary from one decoding iteration to another. 
    The idea is to predict fewer tokens when the model is unsure of its prediction, by assuming that it could be better at the next decoding iteration, for which it will benefit from more context. 
    The first token head is always used to avoid an infinite loop; the other $m-1$ heads are then considered sequentially. 
    If the confidence score (\ie{}, the maximum value after softmax operation) associated with the current head is greater than or equal to $\tau$, then the prediction is preserved, and the next head is processed; the decoding iteration ends otherwise. 

\begin{figure}[ht!]
    \centering
    \begin{subfigure}[c]{\textwidth}
        \includegraphics[width=\linewidth]{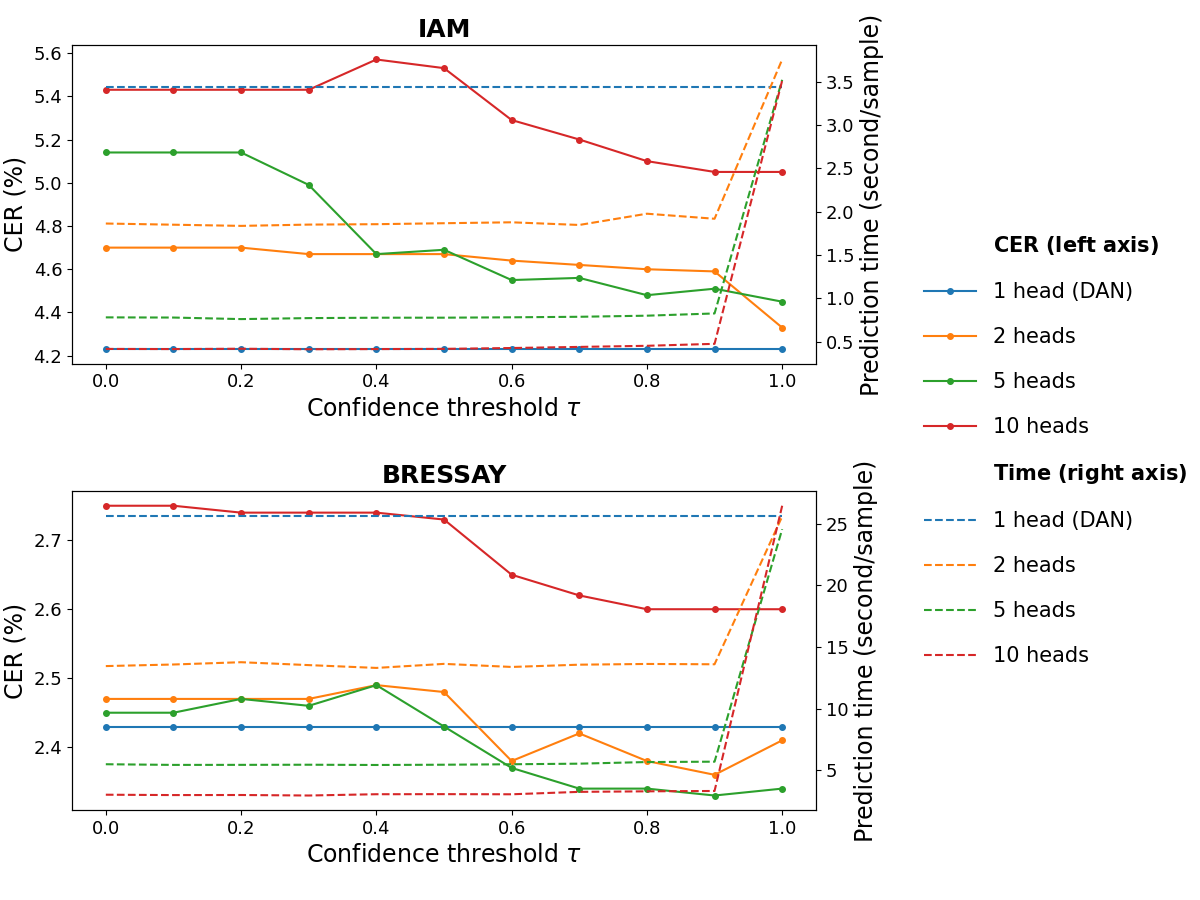}
    \end{subfigure}    
    \caption{Accuracy-speed trade-off with respect to confidence threshold $\tau$ for dynamic decoding of the \mtdans{} on IAM and BRESSAY.} 
    \label{fig:mt-dan-condifence-threshold} 
\end{figure}

We first study the choice of the confidence threshold $\tau$ for the dynamic strategy. Figure~\ref{fig:mt-dan-condifence-threshold} shows the evolution of the CER and of the prediction time with respect to $\tau$ for values between 0 and 1, with a step of 0.1. A confidence threshold of $\tau=0$ means that the token predicted by each projection head is always kept (the fastest option): it corresponds to the static prediction policy. A confidence threshold of $\tau=1$ means that only the prediction of the first projection head is kept (the longest option): it results in the DAN prediction strategy. 
We compare the accuracy-speed trade-off for different numbers of heads on both IAM and BRESSAY datasets. We also show results for a single projection head as a baseline, leading to a CER plateau as the dynamic strategy is designed for multi-token predictions.

It appears that the confidence threshold fulfills its purpose, as there is a trend toward a decrease in the CER as $\tau$ increases, for both datasets. In addition, the prediction time per sample is rather stable from $\tau=0$ to $\tau=0.9$. However, one can note a huge prediction time gap going from $\tau=0.9$ up to $\tau=1$, making the approach ineffective as it consists in single-token prediction, as the DAN.
In this way, we consider $\tau=0.9$ as a good compromise over all the evaluated configurations and use it for the next experiments.

We compare the two prediction policies, namely static and dynamic, in Table~\ref{tab:mt-dan-dec-strategy}. 
For the static prediction policy, we evaluate the models for $k \in \{1, 2, 3, 4, 5\}$ when possible. 
Regarding the IAM dataset, one can observe a trend whereby the CER decreases as $k$ also decreases, with optimum performance reached for $k=1$, noting that decreasing $k$ highly increases the prediction time. 
This trend is less noticeable on the BRESSAY dataset with more fluctuating results. 
However, the proposed dynamic prediction policy with $\tau=0.9$ shows interesting results, which generally outperform the corresponding model using the static strategy with $k=m$, with a limited prediction time overhead.
For instance, it enables to reduce the CER from 5.14\% to 4.51\% for the IAM dataset ($m=5$), while slightly increasing the prediction time from 0.75 seconds (Table~\ref{tab:mtw-dan-results}(b)) to 0.83 seconds.

\begin{table}[ht!]
    \centering
    \caption{Comparison of static and dynamic prediction policies of the \mtdans{}. CER (\%) are given for IAM and BRESSAY datasets. Best results are shown in bold; results that are very close to the best (whose difference in values is less than 0.05\%) are underlined.}
    \resizebox{\linewidth}{!}{
    \begin{tabular}{|c|c|c|c|c|c|c|c|c|c|c|c|c|c|c|}
    \hline
    Number & \multicolumn{7}{|c|}{IAM} & \multicolumn{7}{|c|}{BRESSAY}\\
    \cline{2-15}
    of heads& \multicolumn{5}{|c|}{static} & \multicolumn{2}{|c|}{dynamic} & \multicolumn{5}{|c|}{static} & \multicolumn{2}{|c|}{dynamic}\\
    \cline{2-15}
     ($m$) & $k=1$ & $k=2$ & $k=3$ & $k=4$ & $k=5$ & $\tau=0.9$ & Time & $k=1$ & $k=2$ & $k=3$ & $k=4$ & $k=5$ & $\tau=0.9$ & Time \\
     \hline
     1 (DAN) & 4.23 & \xmark & \xmark & \xmark & \xmark & \xmark & \xmark 
             & 2.43 & \xmark & \xmark & \xmark & \xmark & \xmark & \xmark \\
     \hline
     2  & \textbf{4.33} & 4.70 & \xmark & \xmark & \xmark & 4.59  & 1.92 
        & \underline{2.41} & 2.47 & \xmark & \xmark & \xmark & \textbf{2.36} & 13.60\\
     3  & \textbf{4.31} & 4.85 & 4.74 & \xmark & \xmark & \textbf{4.31} & 1.32
        & \textbf{2.46} & \textbf{2.46} & \textbf{2.46} & \xmark & \xmark & 2.69 & 9.21\\
     4  & \textbf{4.48} & 4.54 & 4.68 & 4.68 & \xmark & \underline{4.52} & 1.00
        & 2.71 & 2.86 & \textbf{2.61} & 2.70 & \xmark & 2.71 & 6.97\\
     5  & \textbf{4.45} & 4.74 & 4.54 & 4.86 & 5.14 & 4.51 & 0.83
        & \underline{2.34} & 2.40 & \underline{2.36} & 2.40 & 2.45 & \textbf{2.33} & 5.70\\
     10 & \textbf{4.79} & \underline{4.80} & 4.87 & 4.87 & 4.90 & \underline{4.80} & 0.47
        & 2.67 & 2.24 & 2.70 & 2.28 & 2.29 & 2.66 & 3.15\\
     25 & 5.73 & 5.61 & 5.58 & 5.64 & 5.68 & \textbf{5.47} & 0.36 
        & 3.35 & 3.30 & 2.92 & \underline{2.91} & 2.93 & \textbf{2.87} & 1.88\\
    \hline
    \end{tabular}
    }
    \label{tab:mt-dan-dec-strategy}
\end{table}

\subsection{\metadans{}}

In this section, we focus on $m=5$ and $w=5$ as it seems to be a good compromise between prediction time and recognition performance. Through experiments, we found out that transfer learning from \mtdans{} to \metadans{} leads to better results than from \wdans{} or line-level pre-training on both IAM and BRESSAY datasets, so we keep this setting in the following.




\paragraph{Comparison with \mtdans{} and \wdans{}}

We compare the \metadans{} with the \mtdans{} and the \wdans{} for the same hyperparameters ($m=5$ and $w=5$) in Table~\ref{tab:meta-dan-comp}. However, it should be noted that this is not a constraint of the proposition, and that $m$ and $w$ can be chosen completely independently for the \metadans{}.
As one can note for the IAM and BRESSAY datasets, combining both windowed queries and multi-token predictions enable to increase the recognition performance, as well as reducing the prediction time.
Compared to DAN, the conclusion is not as clear-cut regarding the character error rate as it is in favor of the DAN for IAM with a CER of 4.23\% compared to 4.79\%, but in favor of the \metadans{} for BRESSAY with a CER of 2.38\% compared to 2.43\%.
It shows that the \metadans{} can be both more efficient and faster than the DAN, depending on the target dataset.

\begin{table}[ht!]
    
\begin{minipage}[c]{0.5\textwidth}

    \centering
    \caption{Comparison of proposed approaches using the static prediction policy ($m=k$).}
    \resizebox{\linewidth}{!}{
    \begin{tabular}{|c|c|c|c|c|c|c|}
    \hline
    \multirow{2}{*}{Approach} & \multirow{2}{*}{$m$} & \multirow{2}{*}{$w$} & \multicolumn{2}{|c|}{IAM} & \multicolumn{2}{|c|}{BRESSAY}\\
    \cline{4-7}
    & & & CER & Time & CER & Time\\
    \hline
    DAN & 1& 1 & \textbf{4.23}& 3.39 & 2.43  & 24.65\\
    \wdans{} & 1 & 5 & 4.87  & 0.77 & 3.02 & 6.63\\
    \mtdans{} & 5 & 1 & 5.14  & 0.75 & 2.45 &  5.28\\
    \metadans{} & 5 & 5 & 4.79 & \textbf{0.43} & \textbf{2.38}  & \textbf{4.01}\\
    \hline
    \end{tabular}
    }
    \label{tab:meta-dan-comp}

\end{minipage}
\hfill
\begin{minipage}[c]{0.45\textwidth}

    \centering
    \caption{Comparison of \metadans{} ($w=5$, $m=5$) performance with respect to the decoding strategy.}
    \resizebox{\linewidth}{!}{
    \begin{tabular}{|c|c|c|c|c|}
    \hline
    Strategy for & \multicolumn{2}{|c|}{IAM} & \multicolumn{2}{|c|}{BRESSAY}\\
    \cline{2-5}
    concurrent & CER  & Time & CER & Time\\
    \hline
    static ($k=1$) & \textbf{4.55} & 0.79 & \textbf{2.31} & 7.16 \\
    static ($k=5$) &  4.79 & \textbf{0.43} & 2.38 &\textbf{ 4.01} \\
    dynamic ($\tau=0.9$) & 4.73 & 0.45 & 2.37 & 7.14\\
    \hline
    \end{tabular}
    }
    \label{tab:meta-dan-dec-strategy}

\end{minipage}
\end{table}

\paragraph{Comparing prediction strategies.}
\label{sec:meta-dan-dec-strat}

As the \metadans{} relies on multi-token predictions, we can apply the same dynamic prediction policy as for the \mtdans{}. 
This is evaluated in Table~\ref{tab:meta-dan-dec-strategy}. 
It appears that the confidence scores associated with the predictions are not as high as those for the \mtdans{}, leading to a failure case for the BRESSAY dataset for which the prediction time is close to that of \wdans{} (static case with $k=1$).
As previously, the static policy with $k=1$ leads to the best recognition performance but suffers from longer prediction.
We consider the static policy with $k=m=5$ as the best alternative for the \metadans{} and use this configuration in the following.

\subsection{Visualization}

\begin{figure}[ht!]
    \centering
    \begin{subfigure}[c]{0.8\linewidth}
        \includegraphics[width=\linewidth]{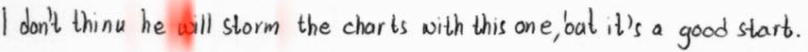}
        \caption{DAN, predicting "w"}
    \end{subfigure}
    ~
    \begin{subfigure}[c]{0.8\linewidth}
        \includegraphics[width=\linewidth]{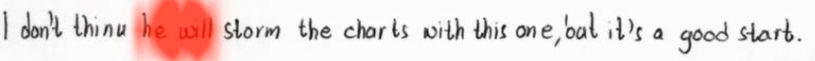}
        \caption{\mtdans{} ($m=5$), predicting "will "}
    \end{subfigure}
    ~
    \begin{subfigure}[c]{0.8\linewidth}
        \includegraphics[width=\linewidth]{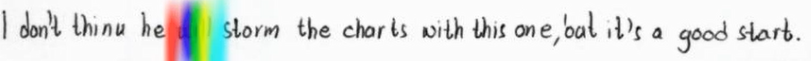}
        \caption{\wdans{} ($w=5$), predicting "will "}
    \end{subfigure}
    ~
    \begin{subfigure}[c]{0.8\linewidth}
        \includegraphics[width=\linewidth]{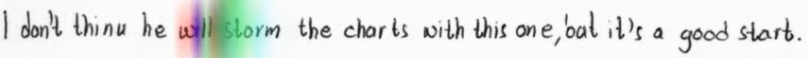}
        \caption{\metadans{} ($m=5$, $w=5$), predicting "will stor"}
    \end{subfigure}
    \caption{Visualization of the attention process for a single decoding step. Attention weights are more diffused when relying on multi-token predictions (\mtdans{}, \metadans{})}
    \label{fig:attention}
\end{figure}

A visualization of the attention process for the different approaches is provided in Figure~\ref{fig:attention}.
Images are cropped to a single line for visibility, but predictions are performed on the full-page images.
Attention weights of the last cross-attention layer are projected on the input image (values are normalized such that the maximum intensity is reached for 0.2). 
One color is associated with each projection head in this order: red, blue, green, yellow, and cyan.
As expected, the attention weights are peakier for DAN and \wdans{} since each query only focuses on a single character prediction, contrary to \mtdans{} and \metadans{} that involve a multi-token prediction strategy.
Interestingly, the attention weights of the \mtdans{} even spread over the last predicted characters ("he ") without disturbing the recognition.
The combination of multi-token predictions and window queries enables the \metadans{} to efficiently predict more tokens (9 here) by taking advantage of more context.

\subsection{Large-scale comparison of the prediction approaches}

\begin{table}[h!]
    \centering
    \caption{CER results (in percentages) on the test sets with FCN-M encoder. We used $m=5$ and $w=5$ for the proposed approaches. Best results at page level are in bold.}
    \resizebox{\linewidth}{!}{
    \begin{tabular}{|c|c|c|c|c|c|c|c|c|c|c|c|c|}
    \hline
    \multirow{3}{*}{Dataset} & \multicolumn{2}{|c|}{Line-level} & \multicolumn{10}{|c|}{Page-level}  \\
    \cline{2-13}
    & \multicolumn{2}{|c|}{FCN} & \multicolumn{2}{|c|}{DAN} & 
    \multicolumn{2}{|c|}{Faster DAN} & \multicolumn{2}{|c|}{\wdans{}} & \multicolumn{2}{|c|}{\mtdans{}} & \multicolumn{2}{|c|}{\metadans{}} \\
    \cline{2-13}
    & CER & Time & CER & Time & CER & Time & CER & Time & CER & Time & CER & Time\\
    \hline
    IAM         & 4.84 & 0.03 & 4.23 & 3.39 & \textbf{4.16} & 0.64 & 4.87 & 0.77 &  4.51 & 0.83 & 4.79 & 0.43 \\
    BRESSAY     & 3.45 & 0.02 & 2.43 & 24.65& 2.34 & 1.51 & 3.02 & 6.63& \textbf{2.33} & 5.70 & 2.38 & 4.01 \\
    READ-2016   & 4.61 & 0.03 & 3.85 & 4.88 & 4.34 & 0.73 & 4.34& 1.09& \textbf{3.80} & 0.80  & 4.22 & 0.63 \\
    Esposalles  & 0.61 & 0.05 & 5.86 & 19.99& \textbf{1.91} &  2.23 & 5.95 & 4.91  & 2.16 & 3.37 & 2.13 & 2.86\\
    ScribbleLens& 4.72 & 0.06 & 4.38 & 16.24  & 4.81 & 2.17 & 5.06 & 4.04   & \textbf{4.16} & 2.71 & 4.31 & 2.24\\
    CASIA-2      & 2.91 & 0.05 & \textbf{1.10} & 2.81  & 3.17 &  0.57& 2.81 & 0.69 & 1.32 & 0.52  & 2.01 & 0.43\\
    Eparchos    & 7.79 & 0.04 & \textbf{7.27} & 9.14     & 7.52 & 0.97  & 8.18 & 2.11  & 8.14 & 1.52 & 7.36 & 1.21\\
    \hline
    RIMES-2009  & \xmark & \xmark & 6.73  & 5.87 & 6.56  & 0.98   & 6.26 & 1.44& 4.94 & 1.09 & \textbf{4.64} & 0.83\\
    RIMES-2024  & \xmark & \xmark & 3.89  & 5.89 & 3.57  & 0.94   & 3.48 & 1.38 &\textbf{2.97} & 1.06 & 3.26 & 0.80\\
    MAURDOR-C3  & \xmark & \xmark & 10.38 & 5.34 & 9.76  &  0.78  & 8.05& 1.27 & 13.71& 1.16 & \textbf{7.93} & 0.69\\
    MAURDOR-C4  & \xmark & \xmark & 8.28  & 6.70 & 10.46 & 1.32  & 6.93& 1.61    & 6.62 & 1.20 & \textbf{6.59} & 0.92\\
    \hline
    Average     & \xmark & \xmark & 5.31 & 9.54 & 5.33 & \textbf{1.17} & 5.36& 2.36& 4.97 & 1.81 & \textbf{4.51} & 1.37\\
    \hline
    \# parameters & \multicolumn{2}{|c|}{17.7 M} & \multicolumn{2}{|c|}{23.1 M} & \multicolumn{2}{|c|}{23.1 M} & \multicolumn{2}{|c|}{23.1 M} & \multicolumn{2}{|c|}{23.4 M} & \multicolumn{2}{|c|}{23.4 M} \\
    \hline
    \end{tabular}
    }
    \label{tab:all-datasets}
\end{table}

We now provide a large evaluation at page level over all datasets and prediction approaches (ours and concurrent ones).
We used the same architecture and training strategy for all the evaluated settings, \ie{}, we also train the concurrent approaches with the same framework for fair comparisons.
We also provide results at line level, which corresponds to the pre-training stage (same encoder with CTC decoding) for comparison.
However, let us recall that line-level metrics only correspond to recognition on humanly-segmented images: it does not include the required segmentation and ordering stages that would lead to higher CER and prediction times. 
Results shown in Table \ref{tab:all-datasets} highlight that it is generally beneficial to process documents at page level rather than line level.
A counter-example is the Esposalles dataset, where line recognition performs very well, demonstrating a certain ease of visual reading (single writer). 
In this particular case, it seems that the benefits of implicit language modeling provided by the transformer decoder are not sufficient to compensate for the added difficulty of learning the reading order. 

Focusing on the prediction time, all proposed approaches are on par with the prediction speed of the Faster DAN. 
However, one can note that the Faster DAN speed improvement is highly dependent on the number of lines to be predicted in the document, contrary to the proposed approaches. 
\textit{E.g.}, the Faster DAN is 16.3 times faster than the DAN for the BRESSAY dataset ($\approx 30$ lines/doc) but 4.9 times faster for the CASIA-2 dataset ($\approx$ 10 lines/doc).

Regarding the CER, the \mtdans{} reaches state-of-the-art performance on 4 datasets. Also, it seems that the enlarged context provided by the \metadans{} stabilizes the attention process, reducing the looping phenomenon (when the same text is predicted multiple times) for heterogeneous layouts as in the MAURDOR datasets.
It results in new state-of-the-art recognition performance when averaging on all datasets with an improvement of 0.46 points compared to \mtdans{} and 0.80 points compared to DAN.
Associated WER results are given in~\ref{sec:appendix-wer} for completeness.

\subsection{Comparison with the state of the art}

\begin{table}[ht!]
    \centering
    \caption{Comparison with state-of-the-art approaches (no external data, no external language model).
    Best result at page level is shown in bold. State-of-the-art line-level results are given for comparison.}
    \resizebox{0.96\linewidth}{!}{
        
    \begin{threeparttable}[b]
    \begin{tabular}{|c|l|c|c|c|c|}
    \hline
    \multirow{3}{*}{Dataset} & \multicolumn{3}{|c|}{\multirow{2}{*}{Approaches of the literature}} & \multicolumn{2}{|c|}{CER (\%) for our}  \\
    & \multicolumn{3}{|c|}{} & \multicolumn{2}{|c|}{page-level approaches}\\
    \cline{2-6}
    & Architecture & Level & CER (\%) &  \mtdans{} & \metadans{}\\
    \hline
    \multirow{3}{*}{BRESSAY} 
        & \cite{BRESSAY} CRNN + attention & line & 3.62 & \multirow{3}{*}{\textbf{2.33}}& \multirow{3}{*}{2.38}\\
        & \cite{BRESSAY} YOLO-v5 + Calamari & line & 6.01 & &\\
        & \cite{BRESSAY} DANIEL & page & 3.55 & &\\
    \hline
    \multirow{2}{*}{CASIA-2} 
        & \cite{Yao2024} CNN + attention & line & 2.34 & \multirow{2}{*}{\textbf{1.32}} & \multirow{2}{*}{2.01}\\
        & \cite{Wang2021} DenseNet + attention & line & 1.78 & &\\
    \hline
    Eparchos 
        & \cite{EPARCHOS} CNN + GRU & line & 4.53 & 8.14 & \textbf{7.36}\\
    \hline
    \multirow{2}{*}{Esposalles} 
        & \cite{Tarride2023b} DAN + NER  & line & 0.48 & \multirow{2}{*}{2.16} & \multirow{2}{*}{\textbf{2.13}}\\
        & \cite{Tarride2023b} DAN + NER  & page & 3.61 & &\\
    \hline
    \multirow{3}{*}{IAM}
        & \cite{Diaz2021} transformer & line & 3.99 & \multirow{3}{*}{4.51} & \multirow{3}{*}{4.79}\\
        & \cite{Coquenet2023b} DAN  & page & 4.54 & &\\
        & \cite{Constum2024} DANIEL  & page & \textbf{4.38} & &\\
    \hline
    \multirow{2}{*}{MAURDOR-C3}
        & \cite{Coquenet2023b} DAN & page & 8.62 & \multirow{2}{*}{13.71} & \multirow{2}{*}{\textbf{7.93}}\\
        & \cite{Coquenet2023c} Faster DAN & page &  8.93 & &\\
    \hline
    \multirow{2}{*}{MAURDOR-C4}
        & \cite{Coquenet2023b} DAN & page & 8.02 & \multirow{2}{*}{6.62} & \multirow{2}{*}{\textbf{6.59}}\\
        & \cite{Coquenet2023c} Faster DAN & page & 9.88 & &\\
    \hline
    \multirow{5}{*}{READ-2016} 
        & \cite{Coquenet2023a} VAN & line & 4.10 & \multirow{5}{*}{3.80} & \multirow{5}{*}{4.22}\\
        & \cite{Coquenet2023b} DAN & page & \textbf{3.43} & &\\
        & \cite{Coquenet2023c} Faster DAN & page & 3.95 & &\\
        & \cite{Constum2024} DANIEL & page &  4.03 & &\\
        
    \hline
    \multirow{3}{*}{RIMES-2009}
        & \cite{Coquenet2023b} DAN  & page & \textbf{4.54} & \multirow{3}{*}{4.94}& \multirow{3}{*}{4.64}\\
        & \cite{Coquenet2023c} Faster DAN  & page & 6.38 & &\\
        & \cite{Constum2024} DANIEL  & page & 5.80 & &\\
    \hline
    RIMES-2024 & \xmark & \xmark & \xmark & \textbf{2.97} & 3.26\\
    \hline
    ScribbleLens & \cite{Dolfing2020} CNN + BLSTM  & line & 6.47 & \textbf{4.16} & 4.31\\
    \hline
    \end{tabular}
    \end{threeparttable}
    }
    \label{tab:sota}
\end{table}

We further compare the proposed approaches to the state of the art in Table~\ref{tab:sota}. 
For fair comparison, we only compare to works following the same constraints, \ie{}, without an external language model and that do not perform training on additional external datasets.
Among the 7 datasets for which there is a page-level evaluation in the literature, we reach new state-of-the-art CER on 4 of these datasets with the \mtdans{} or the \metadans{}. 
The proposed approaches also outperform state-of-the-art line-level recognition models on ScribbleLens and CASIA-2 datasets. 
One can note an exception with the Eparchos dataset, which follows a similar behavior compared to the Esposalles dataset, \ie{}, text recognition is better at line level than at page level. 
This can be explained by the very few available training data that prevents from fully benefiting from the potential of attention-based models.

\subsection{Error analysis}
\begin{figure}
    \centering
    \begin{subfigure}{0.45\textwidth}
        \centering
        \includegraphics[width=\textwidth]{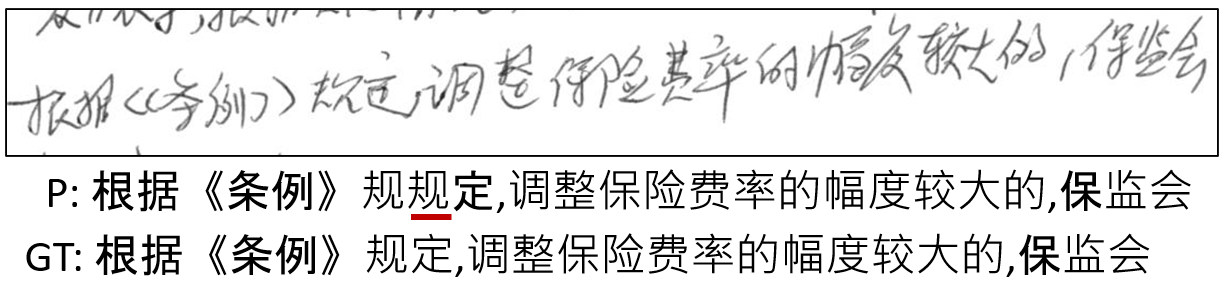}
        \caption{CASIA-2}
        \label{fig:errors-casia-2}
    \end{subfigure}
    \hfill    
    \begin{subfigure}{0.52\textwidth}
        \centering
        \includegraphics[width=\textwidth]{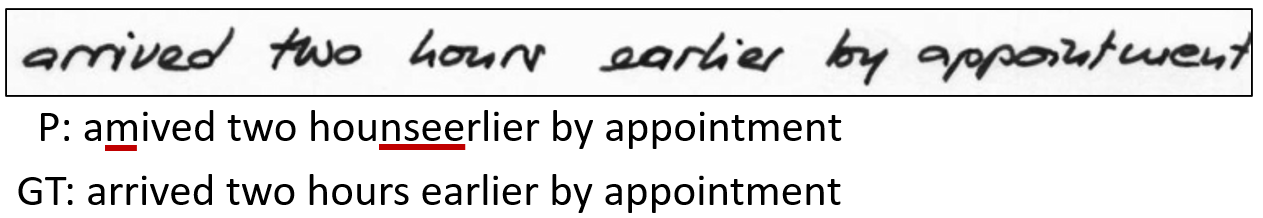}
        \caption{IAM}
        \label{fig:errors-iam}
    \end{subfigure}

    \vspace{0.5cm}

    \begin{subfigure}{0.55\textwidth}
        \centering
        \includegraphics[width=\textwidth]{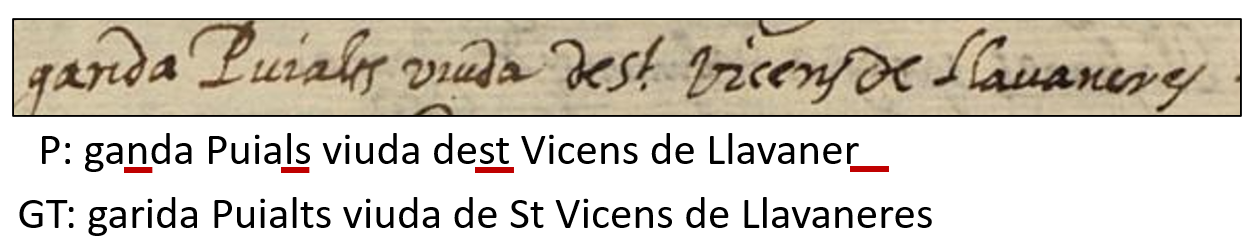}
        \caption{Esposalles}
        \label{fig:errors-esposalles}
    \end{subfigure}
    \hfill    
    \begin{subfigure}{0.4\textwidth}
        \centering
        \includegraphics[width=0.85\textwidth]{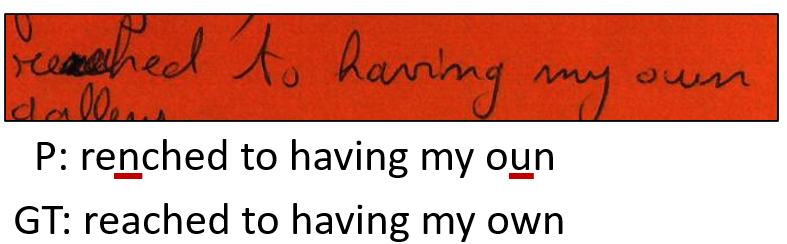}
        \caption{MAURDOR-C3}
        \label{fig:errors-maurdor-c3}
    \end{subfigure}
    
    \vspace{0.5cm}
    
    \begin{subfigure}{0.7\textwidth}       
        \centering
        \includegraphics[width=\textwidth]{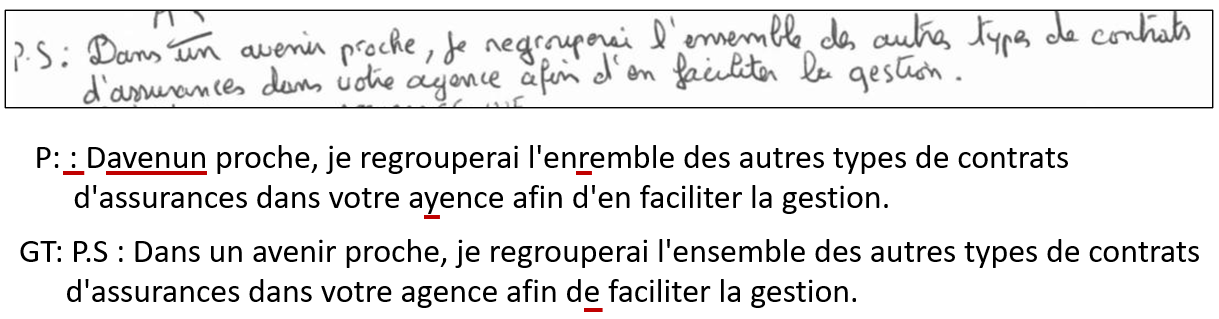}
        \caption{RIMES-2009}
        \label{fig:errors-rimes-2009}
    \end{subfigure}
    
    \caption{Visualization of errors made by \metadans{} on test samples of various datasets. For visibility, page images were cropped and associated with Ground Truth (GT) and Prediction (P). Errors are underlined in red.}
    \label{fig:errors}

\end{figure}

In this section, we propose a qualitative analysis of the errors made by the \metadans{}.
To this end, some examples of predictions and associated ground truths are shown in Figure~\ref{fig:errors} to illustrate the different kinds of errors that can occur.

The most common error is linked to visual ambiguities. For instance, the \metadans{} predicted an "m" instead of "rr" for the word "arrived" (Figure~\ref{fig:errors-iam}) or and "n" instead of an "r" in the word "garida" (Figure~\ref{fig:errors-esposalles}). This shows that, although the proposed contributions improve context modeling, it remains limited: the model sometimes tends to rely more on visual characteristics than on linguistic context. We assume that this is due to the lack of training data in the datasets that prevents the model from really "understand" the text it predicts, leading to the prediction of "dest" instead of "de St" (Figure~\ref{fig:errors-esposalles}), due to the spacing between those two words that is not clearly visible in the image. In other words, it suffers from a lack of linguistic knowledge, but it also lacks knowledge of the world around us to clear up certain ambiguities.

Another kind of error comes from the attention-based prediction process itself. In this way, the same character might be predicted twice (Figure~\ref{fig:errors-casia-2}), or some characters are missed, as in Figure~\ref{fig:errors-rimes-2009} ("P.S"), generally due to specific layouts not seen during training. In rare cases, some characters may have been predicted in the wrong order and partially missed, as "Davenun" predicted instead of "Dans un avenir" (Figure~\ref{fig:errors-rimes-2009}). As mentioned previously, another failure case due to the attention mechanism is related to loops, when the model predicts the same part of text, generally a whole text line, multiple times.

Although rare, predictions may be considered false, wrongly, when the ground truth contains errors as in Figure~\ref{fig:errors-rimes-2009}: the \metadans{} predicted "d'en", as visible in the input image, but the ground truth is "de".

\section{Discussion}

Through this extensive study on prediction strategies, we showed the robustness of the proposed \metadans{} on many HTR dataset configurations, reaching on-average state-of-the-art results.
However, one can note that there is no one strategy that is better on all datasets. 
We noticed that, no matter the prediction variant, it seems to have a beneficial impact on the loop effects, noticeable through the CER results of the Esposalles dataset (Table~\ref{tab:all-datasets}). 
We assume that for the Faster DAN, this is due to the parallelization of the line recognition, since loops mainly occur when switching from one line to another.
For the proposed approaches, this must be the result of the increased context modeling and the forced anticipation of the model to predict the following characters, which can be found on two consecutive lines.
But contrary to the Faster DAN, the \metadans{} does not suffer from "catastrophic" errors. 
Indeed, any missed or duplicated character in the first decoding stage of the Faster DAN leads to discarding or duplicating a whole text line. 
While rare, this can severely impact the recognition performance. 

Regarding the prediction time, it has to be noted that the \metadans{}, while competitive, does not outperform the Faster DAN in this respect, on average. Indeed, although the hyperparameters $m$ and $w$ can be arbitrarily chosen, the experiments showed that in practice, they must be set to a rather low value to preserve competitive recognition performance. Thus, the current state-of-the-art prediction speed for end-to-end page-level text recognition remains an issue for real-time applications.
Although we limit ourselves to comparing the approaches for character-level predictions to reach state-of-the-art results with few training data, all approaches could be combined with BPE tokenization to further reduce the prediction time.

\section{Conclusion}
In this paper, we proposed two new prediction strategies for end-to-end full-page text recognition: windowed queries and multi-token predictions. 
We show that the \metadan{} approach, combining both strategies, reaches new on-average state-of-the-art results on many handwritten datasets while reaching competitive prediction times.
This approach offers a flexible balance between prediction time and recognition performance by adapting the associated hyperparameters to the given use case.
While we only evaluated the \metadans{} for HTR, the proposed approach could be applied to any autoregressive transformer-based system.

However, there is still room for improvement, and some issues remain to be solved regarding end-to-end full-page text recognition. The evaluation protocol of state-of-the-art approaches is based on training a specific model on a single dataset at a time. How to scale them to handle multilingual content, multiple reading orders (mixing Latin and Arabic languages, for instance), and highly heterogeneous layouts remains to be explored. 
These limited training configurations prevent the models from acquiring large, generic linguistic knowledge, leading to some failure cases when the visual content is degraded or ambiguous. Another point is about the prediction time, which remains an issue for real-time applications, as it is still in the order of magnitude of one second per document. Finally, current state-of-the-art approaches consider a single reading order per document, but this is not realistic for very complex layouts (\eg{}, schemas, maps), for which several reading orders could be considered as humanly correct. Future works should focus on adapting metrics and losses to allow multiple reading orders for those cases.

\section*{Acknowledgment}

This work was granted access to the HPC resources of IDRIS under the allocation 2024-AD011012155R2 made by GENCI. Experiments were also carried out using the Grid'5000 testbed, supported by a scientific interest group hosted by Inria and including CNRS, RENATER and several Universities as well as other organizations (see https://www.grid5000.fr).

\bibliographystyle{elsarticle-num} 
\bibliography{references}



\newpage
\appendix
\section{Datasets}
\label{sec-appendix-dataset}

\subsection*{Example images}
\begin{figure}[ht!]

    \begin{subfigure}[c]{0.48\textwidth}
        \includegraphics[width=0.45\linewidth,frame]{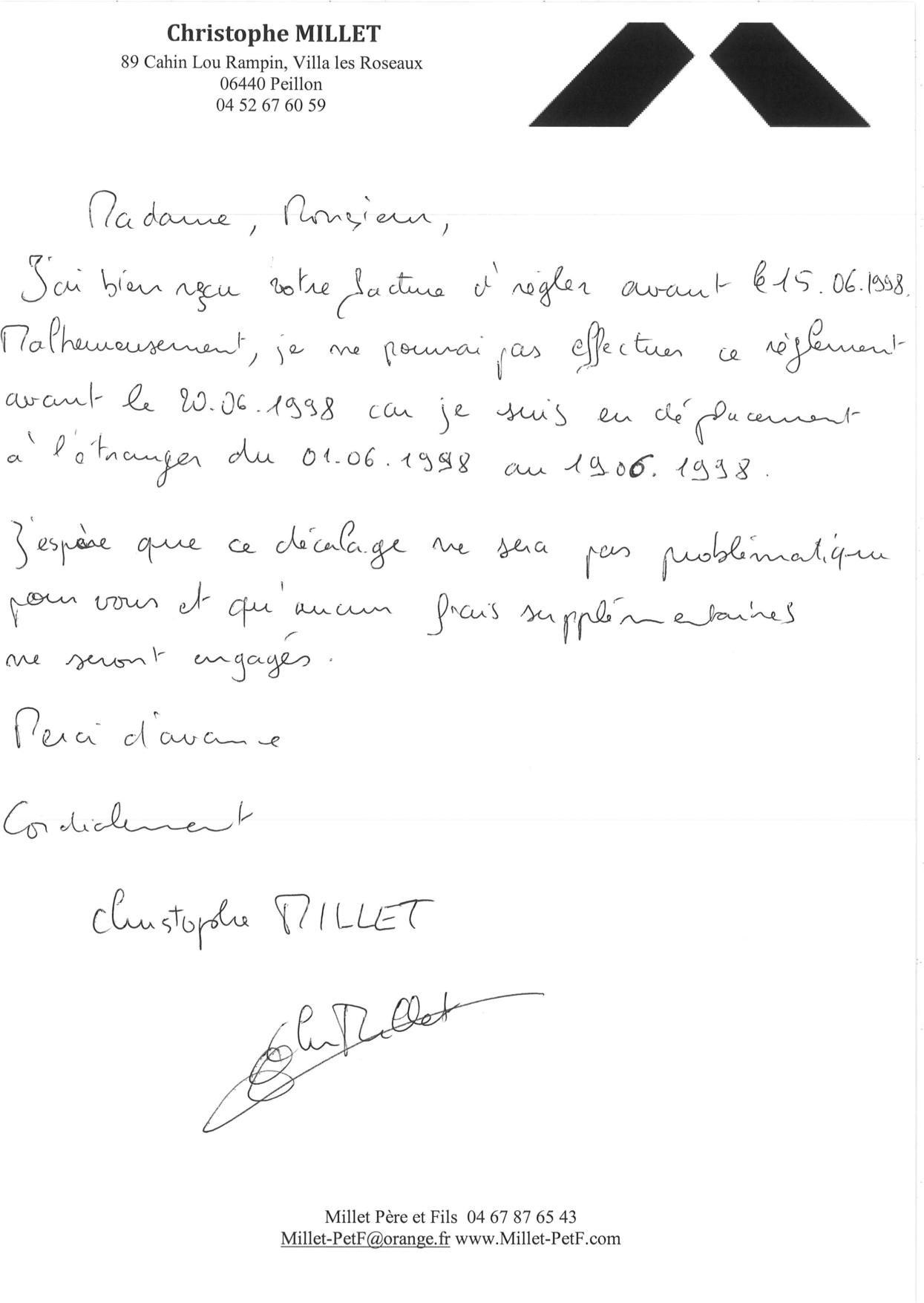}
        \hfill
        \includegraphics[width=0.45\linewidth,frame]{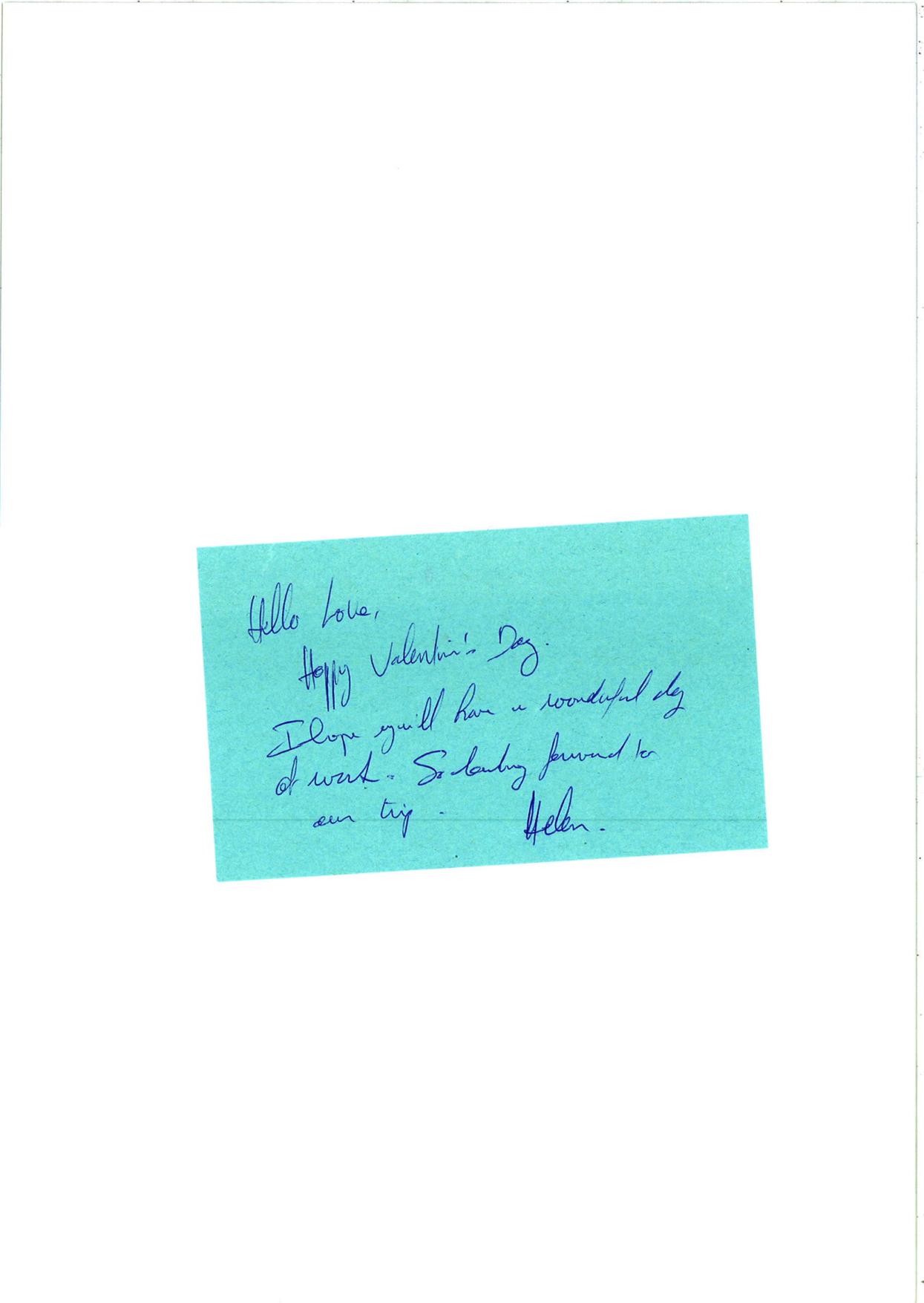}
        \caption{MAURDOR-C3}
    \end{subfigure}
    \hfill
    \begin{subfigure}[c]{0.48\textwidth}
        \includegraphics[width=0.45\linewidth,frame]{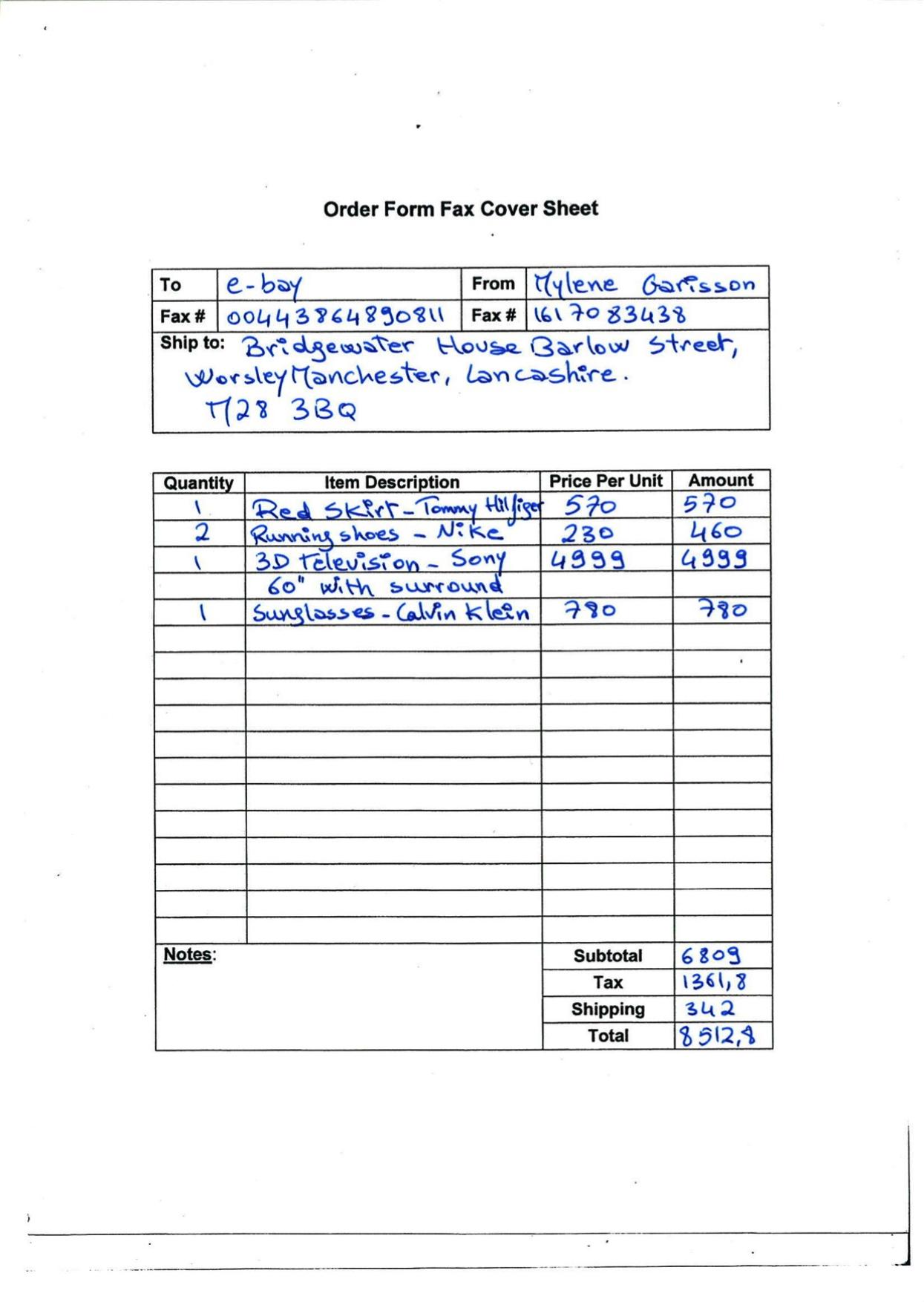}
        \hfill
        \includegraphics[width=0.45\linewidth,frame]{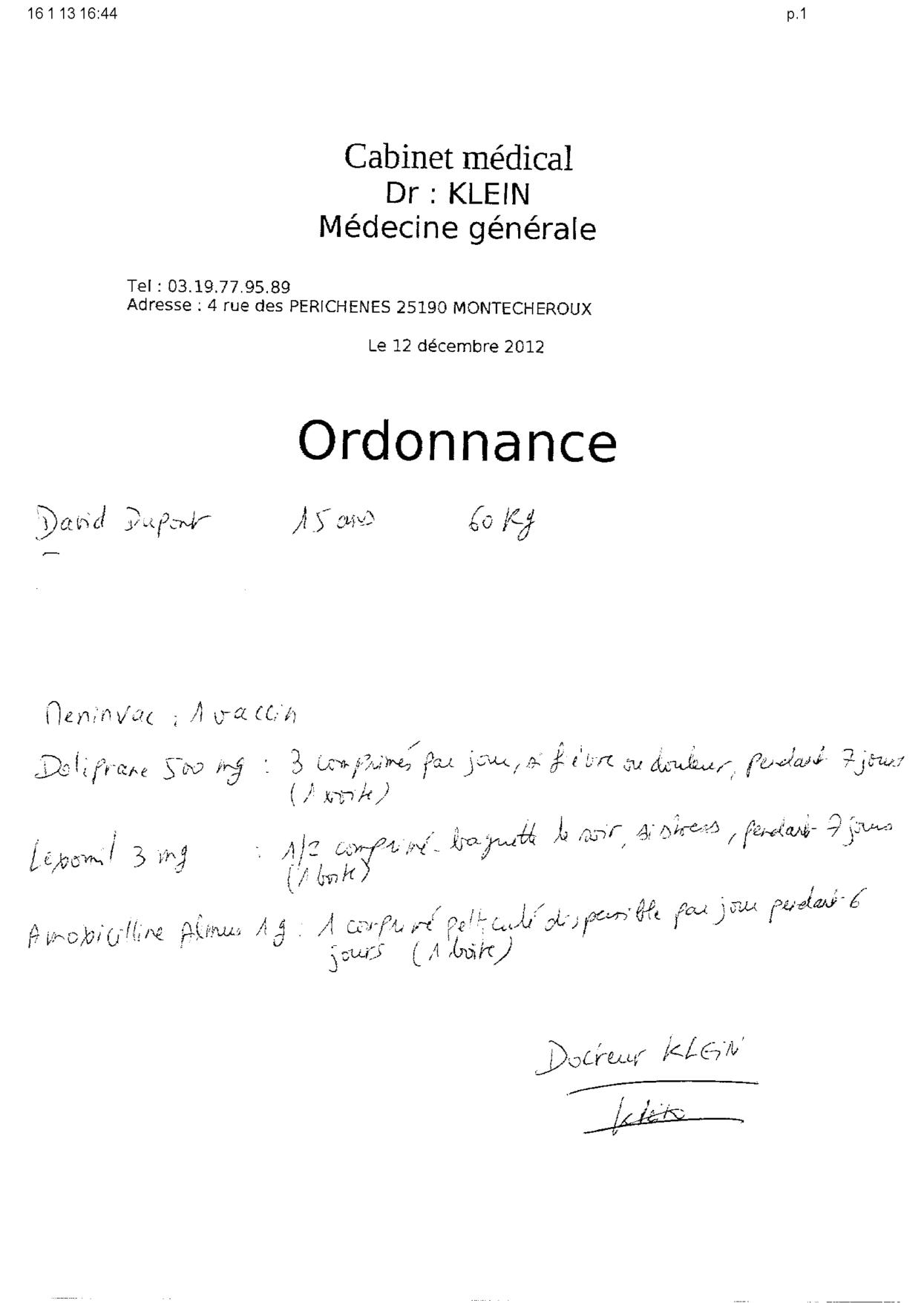}
        \caption{MAURDOR-C4}
    \end{subfigure}

    \vspace{0.2cm}

    \begin{subfigure}[c]{0.22\textwidth}
        \includegraphics[width=\linewidth,frame]{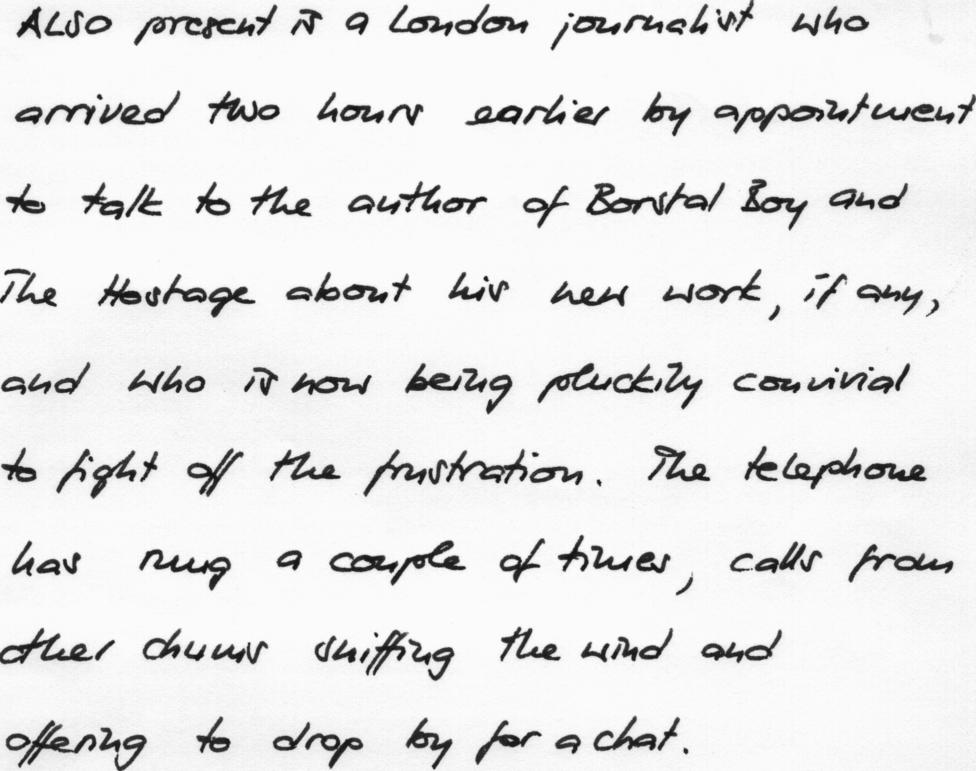}
        \caption{IAM}
    \end{subfigure}
    \hfill
    \begin{subfigure}[c]{0.22\textwidth}
        \includegraphics[width=\linewidth,frame]{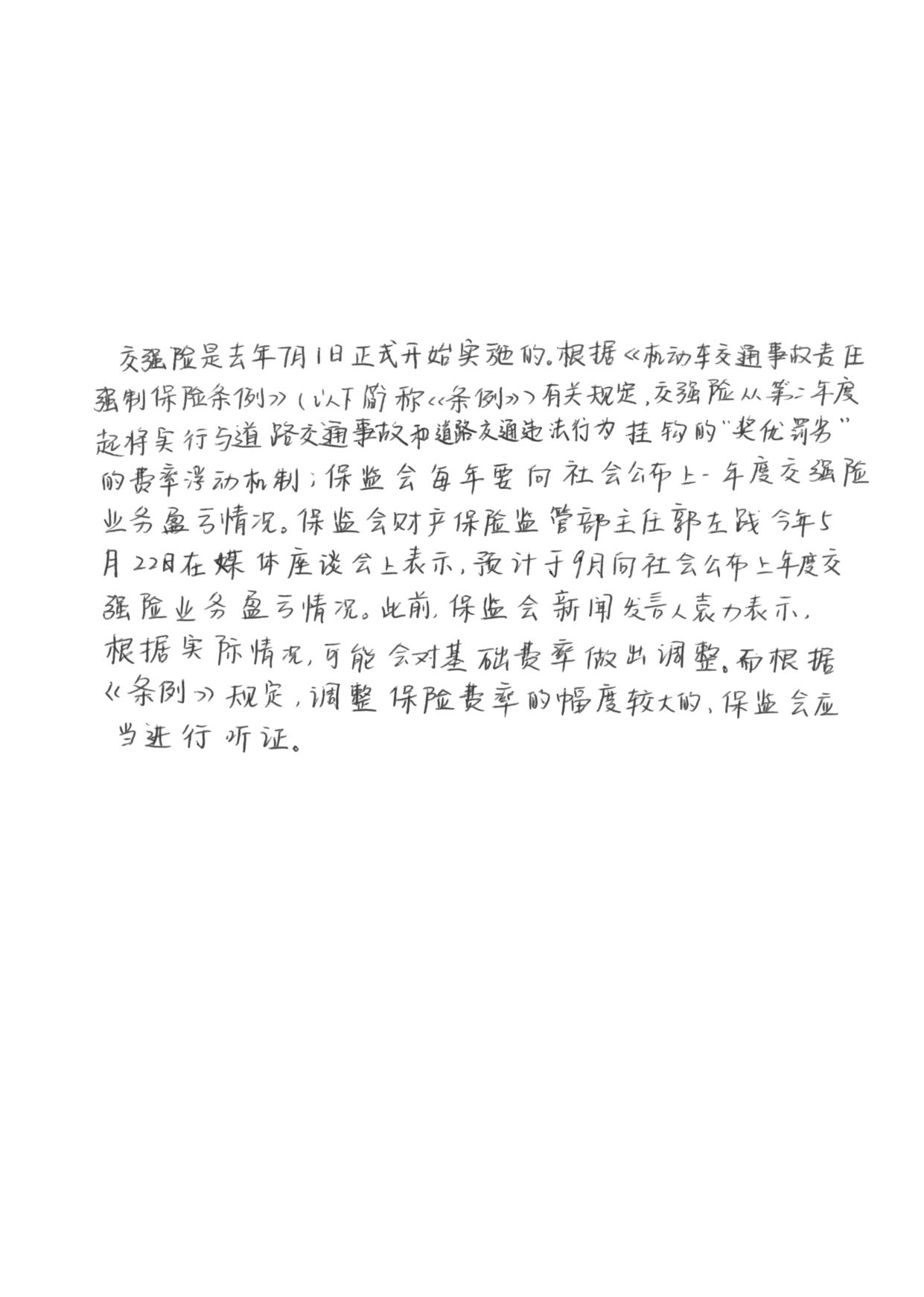}
        \caption{CASIA-2}
    \end{subfigure}
    \hfill
    \begin{subfigure}[c]{0.22\textwidth}
        \includegraphics[width=\linewidth,frame]{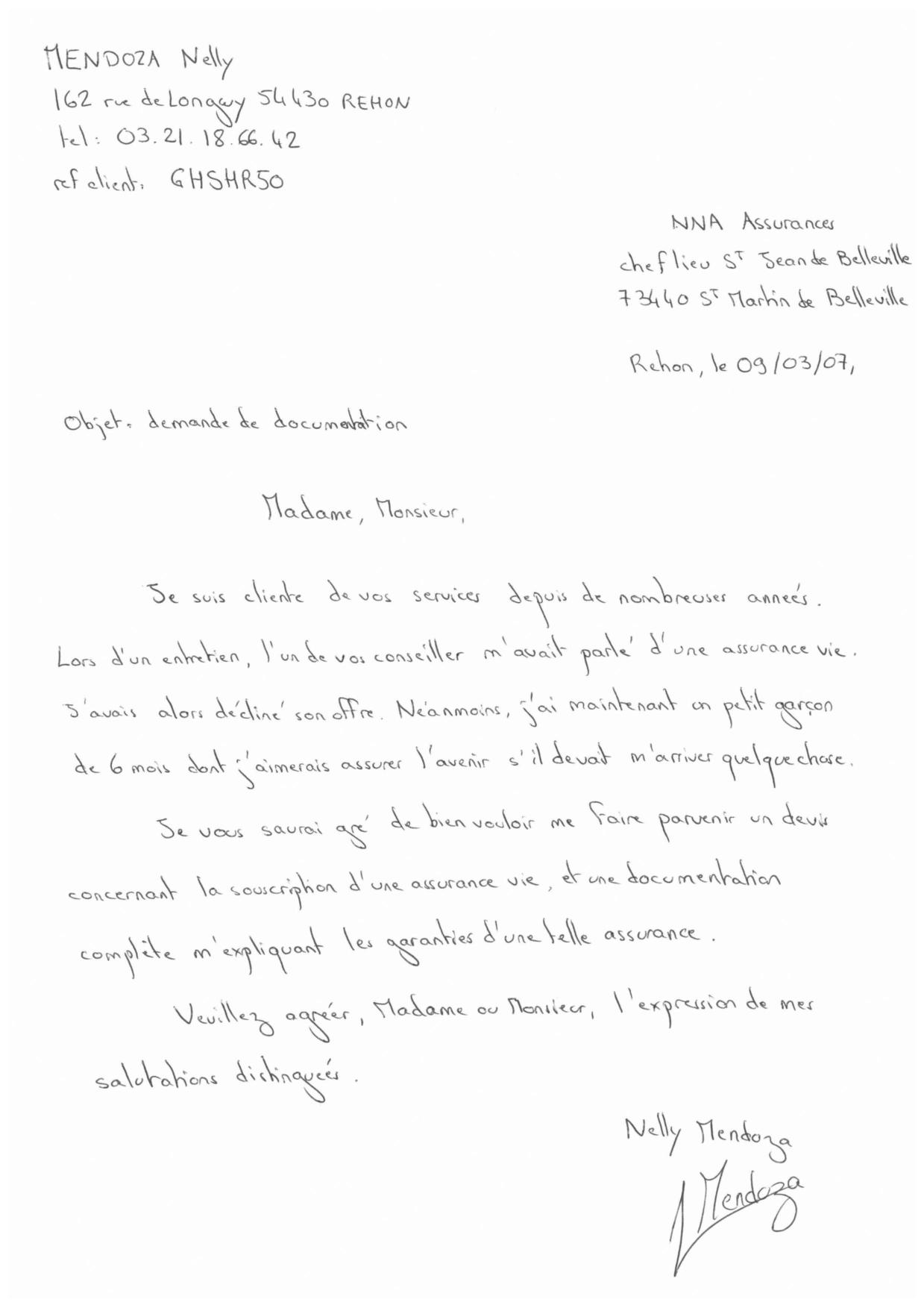}
        \caption{RIMES-2024}
    \end{subfigure}
    \hfill
    \begin{subfigure}[c]{0.22\textwidth}
        \includegraphics[width=\linewidth,frame]{bressay.jpg}
        \caption{BRESSAY}
    \end{subfigure}

    \vspace{0.2cm}

    \begin{subfigure}[c]{0.22\textwidth}
        \includegraphics[width=\linewidth,frame]{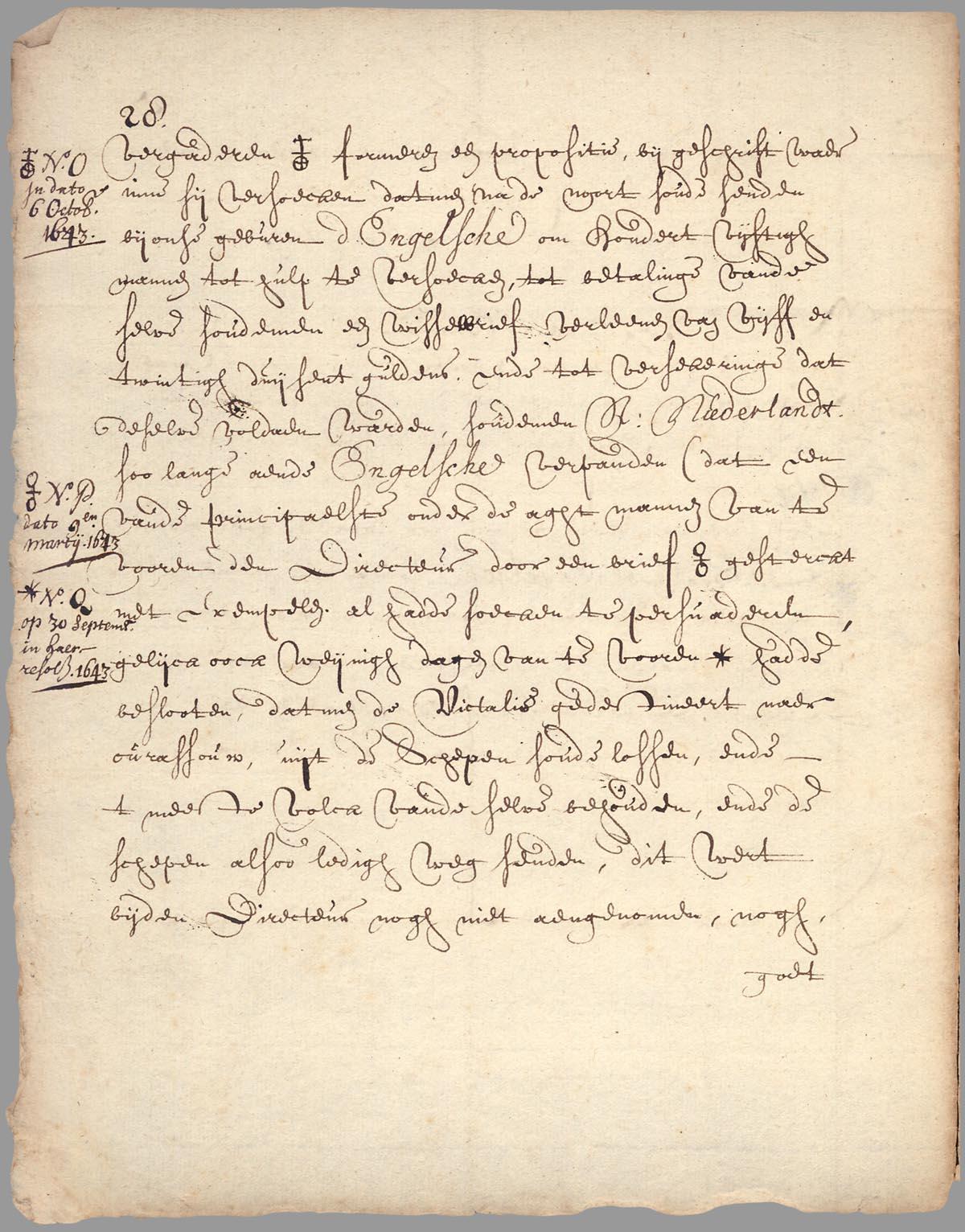}
        \caption{ScribbleLens}
    \end{subfigure}
    \hfill
    \begin{subfigure}[c]{0.22\textwidth}
        \includegraphics[width=\linewidth,frame]{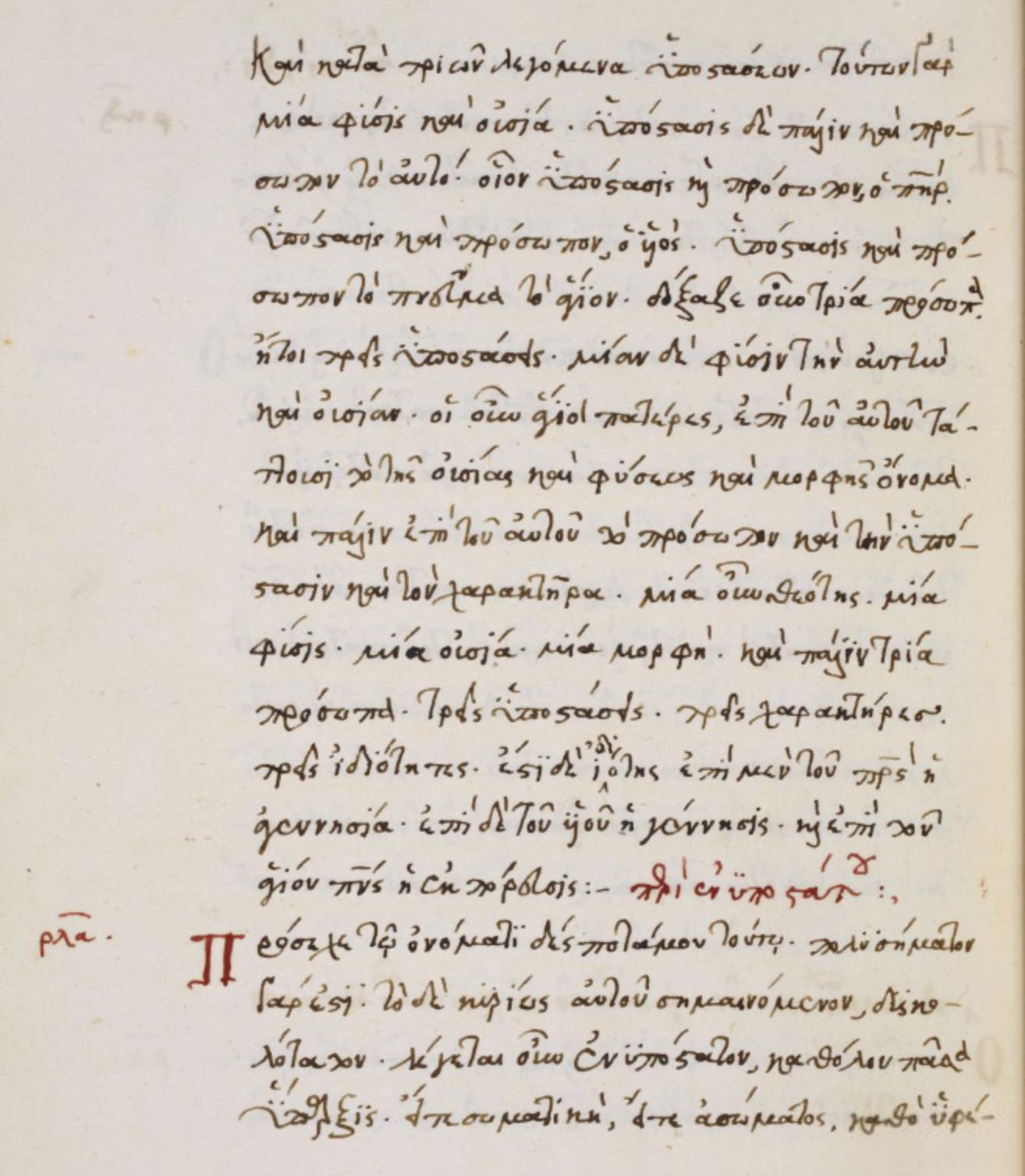}
        \caption{Eparchos}
    \end{subfigure}
    \hfill
    \begin{subfigure}[c]{0.22\textwidth}
        \includegraphics[width=\linewidth,frame]{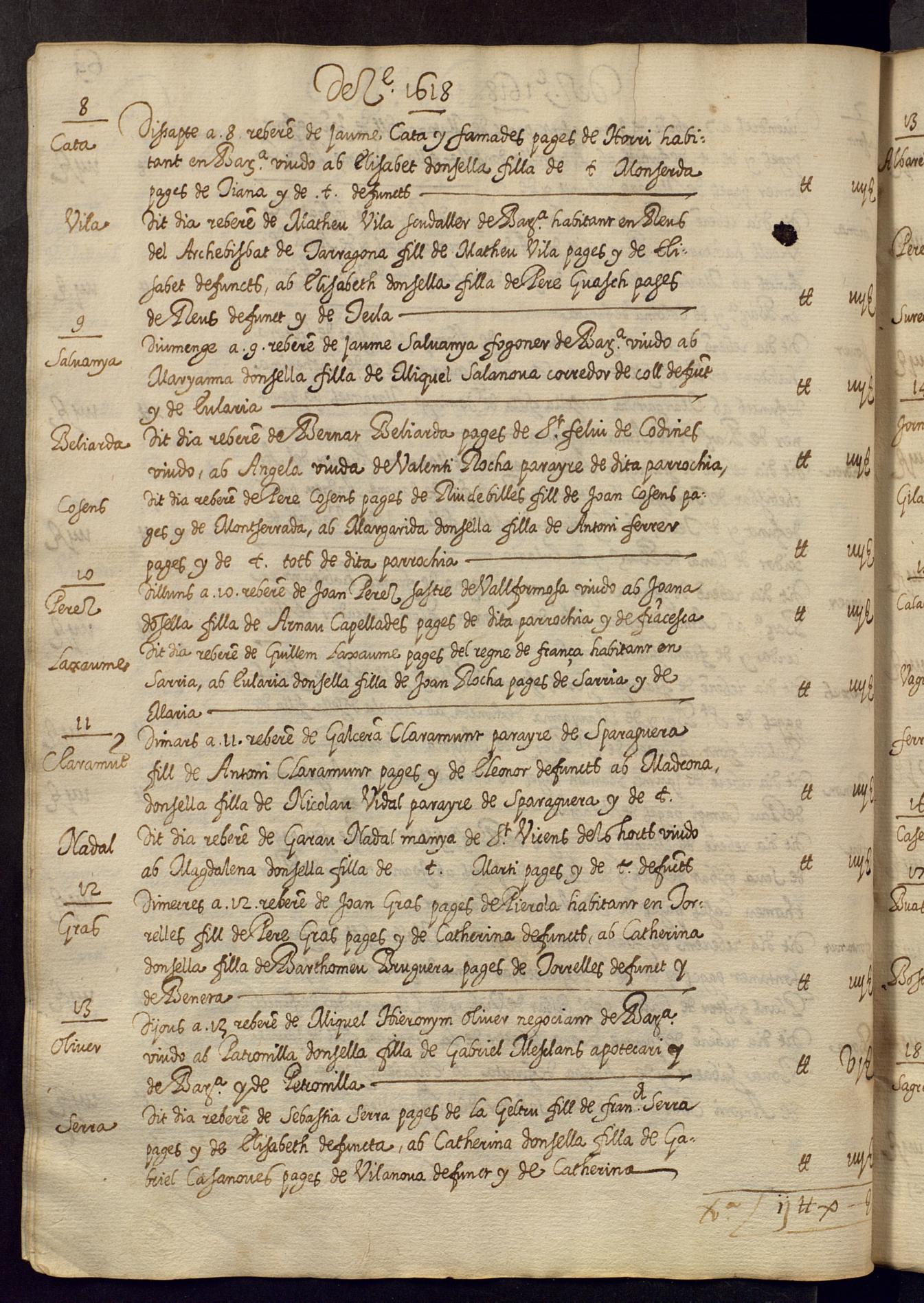}
        \caption{Esposalles}
    \end{subfigure}
    \hfill
    \begin{subfigure}[c]{0.22\textwidth}
        \includegraphics[width=\linewidth,frame]{read2016.jpg}
        \caption{READ-2016}
    \end{subfigure}
    
    \caption{Dataset samples from test sets. The MAURDOR-C3 and MAURDOR-C4 datasets provide widely heterogeneous documents in terms of layout, while the other datasets are quite homogeneous.}
    \label{fig:example-datasets}
\end{figure}

Figure~\ref{fig:example-datasets} provides document image examples from the 10 evaluated datasets. 
As one can note, these datasets cover many page layouts, visual aspects, writing styles, periods, and languages.

\subsection*{Line-level splits}
Table \ref{tab:dataset-line} details the splits for training, validation, and test used for the different datasets regarding the experiments at line level.

\begin{table}[h!]
    \centering
    \caption{Datasets splits at line level.}
    \resizebox{0.5\linewidth}{!}{
    \begin{tabular}{|c|c|c|c|c|c|c|c|}
    \hline
    Dataset & Train & Validation & Test\\
    \hline
        IAM & 6,482 & 976 & 2,915\\
        READ-2016 & 8,367 & 1,043 & 1,140 \\
        Esposalles & 2,328 & 742 & 757 \\
        ScribbleLens & 4,408 & 375 & 563\\
        CASIA-2 & 36,305 & 5,465 & 10,449 \\
        Eparchos & 1,512 & 380 & 380\\
        BRESSAY & 19,561 & 4,608 & 5,915\\
    \hline
    \end{tabular}
    }
    \label{tab:dataset-line}
\end{table}

\section{Scaling up the DAN encoder}
\label{sec:appendix-encoder}
We noticed that the DAN lacked some feature extraction capability when evaluated on challenging datasets such as BRESSAY, which presents some crossed-out text with annotations. 
We opt for scaling up the DAN encoder only in order to increase the performance while preventing a significant prediction time increase. 
Indeed, the encoding part of the model is only used once at inference (it is excluded from the iterative decoding stage); so the computations involved are not such significant compared to decoding. 
The proposed encoder is referred to as FCN-M, and the original one as FCN-S.

\begin{table}[h!]
    \centering
    \caption{Composition of the two versions of the encoder. ConvBlock(X) and DSCBlock(X) indicate the number of feature maps X returned by each convolutional layer.}
    \renewcommand{\arraystretch}{1.5}
    \resizebox{0.6\textwidth}{!}{
        \begin{tabular}{|c|c|c|c|}
            \hline
            & FCN-S \cite{Coquenet2023a} & FCN-M (ours) &  Output size\\
            \hline
            \multirow{10}{*}{\rotate{layers}} & ConvBlock(16) &  ConvBlock(32) & $H \times W $ \\
            \cline{2-4}
            & ConvBlock(32) & ConvBlock(64)  & $\frac{H}{2} \times \frac{W}{2}$ \\
            \cline{2-4}
            & ConvBlock(64) & ConvBlock(128) & $\frac{H}{4} \times \frac{W}{4}$ \\
            \cline{2-4}
            & ConvBlock(128) & ConvBlock(256)  & $\frac{H}{8} \times \frac{W}{8}$ \\
            \cline{2-4}
            & ConvBlock(128) & ConvBlock(512)  & $\frac{H}{16} \times \frac{W}{8}$ \\
            \cline{2-4}
            & ConvBlock(128) & ConvBlock(512)  & $\frac{H}{32} \times \frac{W}{8}$ \\
            \cline{2-4}
            & DSCBlock(128) & DSCBlock(512)  & $\frac{H}{32} \times \frac{W}{8}$ \\
            \cline{2-4}
            & DSCBlock(128) & DSCBlock(512)  & $\frac{H}{32} \times \frac{W}{8}$ \\
            \cline{2-4}
            & DSCBlock(128) & DSCBlock(512)  & $\frac{H}{32} \times \frac{W}{8}$ \\
            \cline{2-4}
            & DSCBlock(256) & DSCBlock(256)  & $\frac{H}{32} \times \frac{W}{8}$ \\
            \hline
            \#  params & 1.7 M & 17.6 M & \\
            \hline
        \end{tabular}
    }
    \label{tab:comparison-encoder-layers}
\end{table}

The proposed modification mainly consists in doubling the number of feature maps per block while preserving the last one, which conditions the size of the decoder, as detailed in Table~\ref{tab:comparison-encoder-layers}. We use the notation of the original paper~\cite{Coquenet2023a}, where ConvBlock is a Convolution Block made up of three convolutions, instance normalization, and dropout. DSCBlock is equivalent except for the convolutions that are replaced by Depthwise Separable Convolutions~\cite{Chollet2017}.

\begin{table}[h!]
    \centering
    \caption{Impact of the encoder changes at line (pre-training stage) and page levels. CER are expressed in percentages, and times in seconds.}
    \resizebox{0.6\linewidth}{!}{
    \begin{tabular}{|c|c|c|c|c|c|}
    \hline
    \multirow{2}{*}{Dataset} & \multirow{2}{*}{Encoder} &\multicolumn{2}{|c|}{FCN (line)} &\multicolumn{2}{|c|}{DAN (page)}\\
    \cline{3-6}
    & & CER& Time& CER& Time \\
    \hline
    \multirow{2}{*}{READ-2016}
    & FCN-S & 4.82 & 0.03 & \textbf{3.82} & 4.61  \\
    \cline{2-6}
    & FCN-M & 4.61 & 0.03 & \underline{3.85} & 4.88 \\
    \hline    
    \multirow{2}{*}{CASIA-2}
    & FCN-S & 5.33 & 0.05 & 1.35 & 2.94\\
    \cline{2-6}
    & FCN-M & 2.91 & 0.05 & \textbf{1.10} & 2.81 \\
    \hline
    \multirow{2}{*}{Bressay}
    & FCN-S & 6.11 & 0.02 & 4.53 & 24.28\\
    \cline{2-6}
    & FCN-M & 3.45 & 0.02 & \textbf{2.43} & 24.65\\
    \hline
    \multirow{2}{*}{IAM}
    & FCN-S & 5.76 & 0.03 & 4.53 & 3.57 \\
    \cline{2-6}
    & FCN-M & 4.84 & 0.03 & \textbf{4.23} & 3.39 \\
    \hline
    \end{tabular}
    }
    \label{tab:encoder-cer}
\end{table}

We evaluate these modifications on READ-2016, CASIA-2, BRESSAY, and IAM datasets at line level, and at page level with the DAN. 
Results are shown in Table~\ref{tab:encoder-cer}. 
As one can note, the modified encoder improves the recognition results (or is equivalent) on all the evaluated datasets, whether it is at line or page level.   
The prediction time is only slightly impacted; we assume this is mainly due to the complexity of learning the reading order. For CASIA-2 and IAM, the prediction time is lower, which can be explained by a reduction of the loop effect (predicting several times the same text part). 
But for READ-2016 and BRESSAY, the prediction time is higher, which can be the result of the additional computations due to the bigger encoder, and to better recognition (missed text parts are now well recognized, adding extra decoding iterations).
The most negative impact is on the READ-2016 dataset with a 6\% increase in the prediction time. 
These results also enhance what has been found in~\cite{Coquenet2023b}: the page-level DAN outperforms its line-level counterpart while they are based on the same encoder.

\section{WER results comparing the prediction strategies.}
\label{sec:appendix-wer}

Table~\ref{tab:wer-all-datasets} provides the WER scores associated with the experiments related to Section VI-E, \ie{}, when relying on the FCN-M encoder.

\begin{table}[h!]
    \centering
    \caption{WER results (in percentages) on the test sets with FCN-M encoder. We used $m=5$ and $w=5$ for the proposed approaches. Best results at page level are shown in bold.}
    \resizebox{0.8\linewidth}{!}{
    \begin{tabular}{|c|c|c|c|c|c|c|}
    \hline
    \multirow{2}{*}{Dataset} & \multicolumn{1}{|c|}{Line-level} & \multicolumn{5}{|c|}{Page-level}  \\
    \cline{2-7}
    & FCN & DAN & Faster DAN & \wdans{} & \mtdans{} & \metadans{} \\
    \hline
    IAM         & 15.83 & 14.00 &\textbf{ 13.42} & 16.28 & 14.96 & 15.78 \\
    BRESSAY     & 12.12 & \textbf{7.85}  & 8.03  & 9.72  & 8.28  & 8.46  \\
    READ-2016   & 20.41 & \textbf{14.69} & 14.83 & 16.90 & 14.90 & 16.03 \\
    Esposalles  & 2.61  & 6.60  & \textbf{2.73}  & 6.63  & 3.09  & 3.09  \\
    ScribbleLens& 20.38 & 19.66 & \textbf{18.64} & 21.19 & 18.92 & 19.71 \\
    Eparchos    & 26.29 & \textbf{22.36} & \textbf{22.36} & 24.71 & 23.39 & 23.69 \\
    \hline
    RIMES-2009  & \xmark& 14.35 & 12.99 & 15.12 & \textbf{12.67} & 12.96 \\
    RIMES-2024  & \xmark& 8.72  & 8.30  & 9.35  & \textbf{8.00}  & 8.72  \\
    MAURDOR-C3  & \xmark& 28.43 & \textbf{19.82} & 20.71 & 27.46 & 20.23 \\
    MAURDOR-C4  & \xmark& 14.80 & 16.19 & 13.50 & \textbf{13.15} & 13.48 \\
    \hline
    Average     & \xmark& 15.15 & \textbf{13.73} & 15.41 & 14.48 & 14.22\\
    \hline
    \end{tabular}
    }
    \label{tab:wer-all-datasets}
\end{table}

Interestingly, the WER results do not follow exactly the same trend as that of the CER, by exhibiting a slight advantage for the Faster DAN, which reaches an on-average WER of 13.73\%, compared to 14.22\% for the \metadans{}.
There is no clear explanation for this observation. However, it still remains lower than the DAN WER, which reaches 15.15\%. 
This would suggest that the Faster-DAN is more reliable when its initial predictions are correct, leading to perfectly recognized words (lower WER), but that bad starts cause cascading errors, leading to higher CER, while the proposed \metadans{} is more constant in character recognition performance.

\end{document}